\documentclass[sigconf, nonacm]{acmart}

\usepackage{algorithm}
\usepackage{algpseudocode}
\usepackage{listings}
\usepackage{float}
\usepackage{enumitem}
\usepackage{comment}

\AtBeginDocument{%
  }

\copyrightyear{2026}
\acmYear{2026}
\setcopyright{rightsretained}
\acmConference[FDG '26]{International Conference on the Foundations of Digital Games}{August 10--13, 2026}{Copenhagen, Denmark}
\acmBooktitle{International Conference on the Foundations of Digital Games (FDG '26), August 10--13, 2026, Copenhagen, Denmark}
\acmPrice{}
\acmDOI{XXXXXXX.XXXXXXX}
\acmISBN{/26/08}

\begin{document}

\title{(Perlin) Noise as AI coordinator}

\author{Kaijie Xu}
\affiliation{%
  \institution{McGill University}
  \city{Montreal}
  \state{Quebec}
  \country{Canada}}
\email{kaijie.xu2@mail.mcgill.ca}

\author{Clark Verbrugge}
\affiliation{%
  \institution{McGill University}
  \city{Montreal}
  \state{Quebec}
  \country{Canada}}
\email{clump@cs.mcgill.ca}

\begin{abstract}
Large scale control of nonplayer agents is central to modern games, while production systems still struggle to balance several competing goals: locally smooth, natural behavior, and globally coordinated variety across space and time. Prior approaches rely on handcrafted rules or purely stochastic triggers, which either converge to mechanical synchrony or devolve into uncorrelated noise that is hard to tune. Continuous noise signals such as Perlin noise are well suited to this gap because they provide spatially and temporally coherent randomness, and they are already widely used for terrain, biomes, and other procedural assets. We adapt these signals for the first time to large scale AI control and present a general framework that treats continuous noise fields as an AI coordinator. The framework combines three layers of control: behavior parameterization for movement at the agent level, action time scheduling for when behaviors start and stop, and spawn or event type and feature generation for what appears and where. We instantiate the framework reproducibly and evaluate Perlin noise as a representative coordinator across multiple maps, scales, and seeds against random, filtered, deterministic, neighborhood constrained, and physics inspired baselines. Experiments show that coordinated noise fields provide stable activation statistics without lockstep, strong spatial coverage and regional balance, better diversity with controllable polarization, and competitive runtime. We hope this work motivates a broader exploration of coordinated noise in game AI as a practical path to combine efficiency, controllability, and quality.
\end{abstract}




\maketitle

\section{Introduction}

Modern open-world and sandbox games must coordinate large numbers of agents and generate vast worlds that feel coherent over space and time. Typical player-facing scenarios include: thousands of pedestrians sharing streets without marching in lock-step, ambient NPCs whose routines plausibly bridge the gaps between authored quests, and encounter/spawn systems that feel fair and varied across a huge map. 
At production scale, developers must simultaneously balance local smoothness, global de-locking, and runtime efficiency. Research and industry practice offer many building blocks: flocking and social-force models for collective motion \cite{reynolds1987flocks,helbing1995social}, field-guided crowds that combine global guidance and local avoidance \cite{treuille2006continuum}, reciprocal velocity-obstacle methods and parallel avoidance \cite{van2008reciprocal,guy2009clearpath}, world-scale navigation via HPA*/JPS/WHCA* \cite{botea2004near,harabor2011online,silver2005cooperative}, navmesh pipelines and guidance fields \cite{van2016comparative,patil2010directing}, as well as scheduling/LOD patterns from production texts \cite{yannakakis2018artificial,wissner2010level}. Open-world case studies further highlight ambient AI architectures for persistence \cite{plch2014aiow}, behavior-object encapsulation to manage complexity \cite{cerny2017behaviorobjects}, and rule-based smart-event systems for controllable triggers \cite{zielinski2021smartevents}. However, these approaches require nontrivial inter-agent communication, centralized planning, or handcrafted scripting to maintain variety at scale; search-based methods can discover diversity \cite{kirk2024diverse} but at high compute cost. A lightweight, seedable, and temporally coherent coordination framework that scales to millions of cheap decisions per second remains highly desirable.

Procedural noise provides smooth, controllable randomness, long used in graphics and games. In particular, it offers multi-octave, low-frequency fields with strong spatial structure and efficient evaluation \cite{perlin1985image,ebert2002texturing,lagae2010state}. Such fields already support terrain/biome pipelines \cite{green2005implementing, geiss2007generating, lagae2009procedural}, and align with hierarchical landscape structure beyond graphics \cite{etherington2022hnlm}. We hypothesize that these low-frequency, temporally coherent random fields are well-suited as a general layer for large-scale game AI coordination. Accordingly, we propose a framework that uses seeded Perlin fields to: (i) parameterize behavior, (ii) modulate activation timing, and (iii) generate world-type feature layers consistent with PCG practice \cite{smelik2014survey, shaker2016procedural, hendrikx2013procedural}.

We adopt classic Perlin noise as the representative coherent-noise primitive because it is ubiquitous, well-characterized, fast, and easy to reproduce \cite{perlin1985image,ebert2002texturing,lagae2010state}; validating strong results with this baseline suggests the broader promise of coherent-field coordination before exploring alternatives (e.g., Simplex or sparse Gabor noise \cite{lagae2009procedural}). For behavior parameterization, we drive agent heading and speed with separate low-frequency fields plus light smoothing; for activation timing, we instantiate two sub-directions: (A) hazard (per-timestep activation-rate) /phase-driven action schedulers with global rate control, and (B) spawn placement via space-time couplings under quotas. For type-feature world generation, we produce discrete layout layers via regional quantile mapping and continuous features via mid/high-frequency transforms. Across baselines in each area, our field-driven approach has more coherent crowds, smoother active-count series at matched duty cycles, and spawn fronts with balanced coverage, all while remaining seedable and efficient. Qualitative results show locally related but globally varied flows, reinforcing coherent noise as a compact substrate for scalable game-AI coordination. We hope this encourages more work on deliberately shaping noise fields as a practical tool for large-scale, efficient AI control. Our contributions are:

\begin{itemize}
\item We present a general Perlin-as-coordinator approach and a practical framework that uses coherent random fields as a lightweight basis for large-scale AI control in games. 

\item We instantiate three directions: behavior parameterization, activation timing, and type-feature world generation, with concrete designs consistent with PCG and AI pipelines. 

\item We provide quantitative and qualitative evaluations against diverse baselines, mapping trade-offs between local coherence, global de-locking, and runtime cost, and showing reproducibility via seeded substreams.
\end{itemize}
\section{Related Works}

\subsection{Large-Scale AI Coordination and Simulation}

Research on collective motion and crowd dynamics provides the algorithmic foundations for coordinating large numbers of agents. Classic work such as Boids and social-force models validates how flocking and pedestrian flows emerge from decentralized local rules and continuous attraction/repulsion dynamics \cite{reynolds1987flocks,helbing1995social}. These ideas were extended with potential/flow-field methods (e.g., Continuum Crowds) \cite{treuille2006continuum}, reciprocal collision avoidance (ORCA) \cite{van2008reciprocal}, and GPU-friendly schemes such as ClearPath \cite{guy2009clearpath}, generating rule-based, field-based, and velocity-obstacle formulations that scale well to large crowds. 
Production pipelines combine guidance fields and steering \cite{patil2010directing} with time-sliced scheduling and AI level-of-detail to meet frame budgets \cite{yannakakis2018artificial,wissner2010level}; ambient-AI case studies describe persistence- and component-based scheduling, behavior-object architectures, and smart-event systems for authorable triggers and propagation \cite{plch2014aiow,cerny2017behaviorobjects,zielinski2021smartevents}. Recent frameworks such as TaiCrowd confirm that careful data layout and batch-parallel updates keep large agent sets interactive on commodity hardware \cite{guan2025taicrowd}. 
We replace explicit neighbor communication and centralized scheduling with seedable, temporally coherent noise fields that act as lightweight control signals, retaining the local coherence of guidance/flow-field approaches while remaining compatible with common steering/LOD pipelines and modern crowd frameworks \cite{treuille2006continuum,patil2010directing,yannakakis2018artificial,wissner2010level,guan2025taicrowd}.

\subsection{Procedural World/Map Generation and Type-Feature Decoupling}

Procedural content generation (PCG) offers abstractions for synthesizing large environments under structural constraints, with surveys distinguishing constructive pipelines from generate-and-test and search/evolutionary approaches \cite{shaker2016procedural,hendrikx2013procedural,smelik2014survey,freiknecht2017survey}. 
Game-focused world and ecosystem PCG typically layers low-frequency fields for macro geography with mid/high-frequency fields and masks for vegetation, rarity, and encounter intensity, often combined with terrain generation and histogram shaping to respect design constraints \cite{smelik2014survey,shaker2016procedural}. Spatial placement enforces coverage and separation via blue-noise and Poisson-disk sampling \cite{bridson2007fast,lagae2008comparison,yan2015survey}.
Beyond terrain, open-world systems coordinate populations and resources: constraint satisfaction injects situation-driven NPC vignettes into live simulations \cite{cerny2014spice}, spawn controllers use statistical placement to regulate fairness and availability \cite{eyniyev2023spawn}, and geo-spatial analyses of location-based games such as \emph{Pokémon Go} reveal biases in points-of-interest and spawn distributions that motivate controllable intensity fields and quotas \cite{juhasz2017pokemon}. 
Case studies further explore world-scale map synthesis via multi-agent logic on noise-shaped overworlds and voxel worlds with rule-based structures such as castles, alongside dungeon-focused generators that emphasize connectivity and layout control \cite{hansson2024overworld,waikato2015voxelcastles,shen2022procedural}. Our framework extends these pipelines by using seedable fields for contiguous discrete layers (faction, biome, danger), separate mid/high-frequency fields for continuous attributes (density, rarity, power), and inhomogeneous Poisson placement with class-wise quotas and radii, turning type-feature decoupling into an explicit, reusable design pattern rather than an implicit side-effect of specialized controllers \cite{bridson2007fast,lagae2008comparison,yan2015survey,cerny2014spice,mitchell2022neveradull,eyniyev2023spawn,aydin2023reputation}.

\subsection{Noise and Random Fields in Games}

\begin{figure*}[!t]
    \centering
    \includegraphics[width=\linewidth]{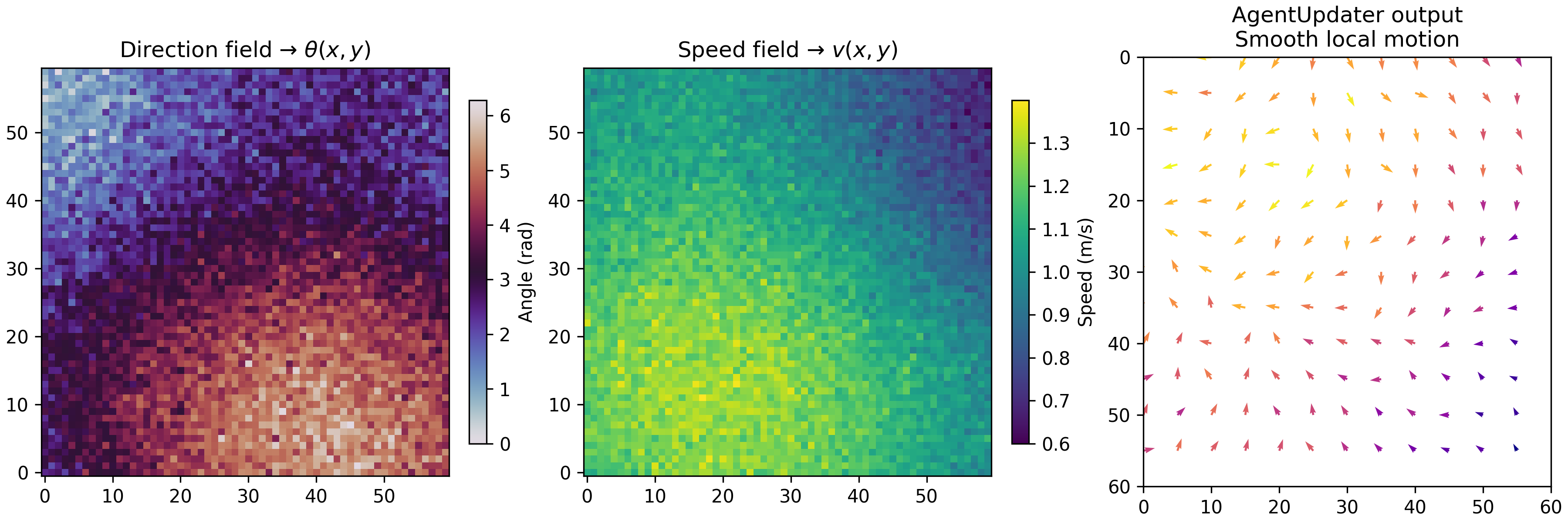}
    \caption{Behavior parameterization. Left: low-frequency heading field $\theta(\mathbf{x}, t)$. Middle: independent speed field $v(\mathbf{x}, t)$. Right: AgentUpdater output after blending previous state with Perlin targets; arrows show local headings and colors encode speeds.}
    \label{fig:behavior_fields}
\end{figure*}

\begin{figure*}[t]
    \centering
    \includegraphics[width=\linewidth]{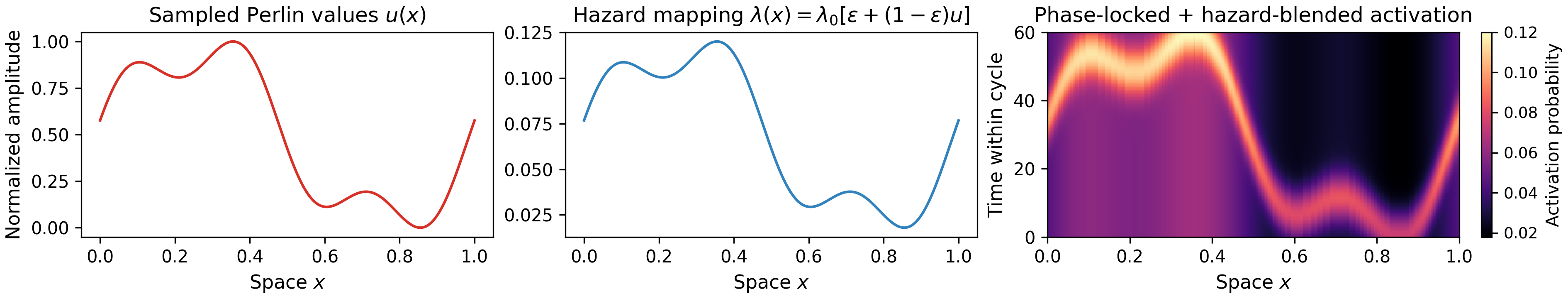}
    \caption{Temporal modulation along a 1D spatial slice of our 2D Perlin field.
    Left: Perlin values $u(x)$ sampled along a line in space.
    Middle: hazard mapping $\lambda(x)=\lambda_0[\varepsilon+(1-\varepsilon)u(x)]$.
    Right: mixed activation probability over cycle time and this spatial coordinate; in practice this mapping lets an agent's position determine when within each cycle it is expected to activate or take actions.
    In the full system the field is defined over $\mathbf{x}\in\Omega\subset\mathbb{R}^2$; we show a 1D slice here for clarity.}
    \label{fig:activation_hazard}
\end{figure*}

Procedural noise provides smooth, controllable randomness that has supported graphics and game content for decades. 
Perlin's gradient noise introduced coherent random fields for efficient synthesis of natural textures and spatial structure \cite{perlin1985image}; multi-octave compositions add scale-rich detail \cite{ebert2002texturing}, and surveys summarize key implementation trade-offs such as frequency control and tiling \cite{lagae2010state}.
Neutral landscape models in ecology further emphasize the alignment between multi-octave structure and large-scale landscape patterns \cite{etherington2022hnlm}. In games, Perlin/fBM pipelines underlie terrain, biome, and resource generation: workflows layer low-frequency fields for macro layout with mid/high-frequency fields for micro-variation and authoring constraints \cite{hendrikx2013procedural,shaker2016procedural,smelik2014survey}, supported by improved Perlin noise, large-scale terrain synthesis, and alternative models such as sparse Gabor noise \cite{green2005implementing,geiss2007generating,lagae2009procedural}. Practical applications include sandbox and exploration maps, Unity/GPU-based 3D world generators, and landscapes that combine Perlin/fBM with erosion and water modeling \cite{indonesia2022perlin,aui2022pg3d,ucu2020landscape}, often pairing noise fields with blue-noise or Poisson-disk processes when strict spatial separation is required \cite{smelik2014survey}. Noise fields also act as control media for interactive behaviors: curl-noise yields divergence-free vector fields for particle and smoke motion \cite{bridson2007curl}, and field-guided approaches use navigation potentials to steer crowds in large environments without explicit agent-to-agent messaging \cite{treuille2006continuum}. PCG literature has long suggested using temporally coherent scalar fields to modulate event and encounter distributions under authorial constraints \cite{hendrikx2013procedural,smelik2014survey}, technically supported by higher-dimensional sampling and slow drift for temporal coherence \cite{lagae2010state,ebert2002texturing}. We treat random fields as a general control substrate: the same Perlin stacks that shape terrain and biomes also drive neutral crowd flows, hazard/phase schedulers, and type-feature layers in our framework. Supplementary control operators turn classic noise primitives into a practical alternative to neighbor-based coordination or hand-scripted spawn logic, producing structured yet globally de-locked space-time patterns compatible with existing PCG and crowd/encounter systems.

\section{Methodology}
\label{sec:method}

This section formalizes a general Perlin-driven framework built around three directions. We present shared modeling and show the instantiations in our experiments.

\subsection{General Perlin Framework}

Let the spatial domain be a square $\Omega\subset\mathbb{R}^2$ of side length $L$ with wrap-around (toroidal, default) or reflective boundaries, and discrete time $t\in\{0,\dots,T-1\}$ with optional cycle length $T_{\mathrm{cycle}}$. Agents $i\in\{1,\dots,N\}$ have position $\mathbf{x}_i(t)\in\Omega$, heading $\theta_i(t)\in[0,2\pi)$, and speed $v_i(t)\ge 0$. We expose a decision context $\mathbf{d}=(x,y,t,c,s_1,\dots,s_k,seed)$, i.e., the information passed to the noise fields when making a decision, where $c\in\mathcal{C}$ indexes a discrete class/identity (e.g., NPC type, faction, event class) and $s_j\in[0,1]$ are normalized state features (e.g., local density, threat, diurnal phase, weather). We maintain a \emph{dual-field} design: $N_{\texttt{type}}(\mathbf{d})$ drives discrete classes/layouts (what appears where), while $N_{\texttt{feat}}(\mathbf{d})$ drives continuous intensities (how strong or dense things are there), with independent seeds to avoid type-strength locking.

A scalar multi-octave Perlin field $n(\mathbf{x},t)$ is constructed by stacking $K$ octaves with base frequency $f$, persistence $p\in(0,1]$, and lacunarity $\ell>1$. Writing $A_K=\sum_{k=0}^{K-1}p^k$ and denoting classic Perlin noise (with fade and lattice gradients) by $P(\cdot)$, we define
\[
\tilde{n}(\mathbf{x},t)=\sum_{k=0}^{K-1} p^k\,P\!\big(\ell^k f\,x+\phi_x,\ \ell^k f\,y+\phi_y,\ t+\phi_t\big),\quad
n(\mathbf{x},t)=\frac{\tilde{n}(\mathbf{x},t)}{A_K},
\]
where offsets $\boldsymbol{\phi}=(\phi_x,\phi_y,\phi_t)$ encode deterministic seeding and temporal phase. We normalize to $[0,1]$ by $n_{01}(\mathbf{x},t)=\tfrac{1}{2}\big(n(\mathbf{x},t)+1\big)$ or via min-max. Two update modes are supported without breaking continuity: \emph{drift}, which advances a global phase $t\leftarrow t+v_{\mathrm{drift}}$ (equivalently translating the sampling window) to advect patterns smoothly; and \emph{resample}, which refreshes offsets at cycle boundaries so that the field is constant within each cycle but can change between cycles, inducing piecewise-stationary regimes.

All downstream mappings reuse the same construction with distinct seeds and octave stacks. Discrete layouts use a regional quantile map $Q$ so that $C(\mathbf{x})=\mathrm{bin}\!\big(Q(N_{\texttt{type}}(\mathbf{x}))\big)$ matches target class proportions while preserving belt contiguity, i.e., we sort values per region and cut them into bands with the desired fractions. Continuous intensities map $N_{\texttt{feat}}(\mathbf{x})$ to parameter ranges, e.g.,
\[
\theta(\mathbf{x},t)=2\pi\,N_{\theta}(\mathbf{x},t)+\zeta,\qquad
v(\mathbf{x},t)=v_{\min} + N_{\nu}(\mathbf{x},t)\,(v_{\max}-v_{\min}),
\]
with a small jitter $\zeta$ for de-locking (breaking up perfectly synchronized, lockstep agent motion). Time-phase is $\,\tau(\mathbf{x})=\lfloor N_{\phi}(\mathbf{x})\,T_{\mathrm{cycle}}\rfloor\,$, yielding locally consistent activations. We map the field to per-timestep activation probabilities with a chosen global average rate:
\[
\lambda(\mathbf{x},t)=\lambda_0\big(\varepsilon+(1-\varepsilon)N_{\lambda}(\mathbf{x},t)\big),\qquad
\Pr[\mathrm{activate}]=1-\exp\!\big(-\lambda\,\Delta t\big),
\]
with $\varepsilon\in[0,1)$ avoiding empty windows; mean normalization rescales $\lambda$ to a target global rate without altering relative spatial structure. All components are reproducible via a seed bundle $\{layout,noise,place\}$ for each layer.

\subsection{Direction I: Behavior Parameterization for Crowd Motion}

Behavior Parameterization provides continuous control fields for agent dynamics while preserving local coherence. Our crowd model targets background, non-goal-directed flow, and is intended as a lightweight layer that can be combined with standard navigation/avoidance if needed. Concretely, we \textbf{sample multiple independent Perlin noise fields at each agent's position to read off} target heading and speed \textbf{parameters}, so that motion is coordinated by the underlying field rather than by explicit agent-agent communication. Given the decision context above, the heading and speed targets follow
\[
\theta(\mathbf{x}, t)=2\pi N_\theta(\mathbf{x}, t)+\zeta,\qquad
v(\mathbf{x}, t)=v_{\min}+N_\nu(\mathbf{x}, t)(v_{\max}-v_{\min}),
\]
with a small jitter $\zeta$ preventing perfect lock-step. Figure~\ref{fig:behavior_fields} shows the resulting flow: the left panel illustrates the heading field, the middle panel shows the companion speed field, and the right panel depicts the smooth AgentUpdater output $\mathbf{v} = v(\cos\theta, \sin\theta)$ after vector averaging and exponential smoothing.
Agents evolve on a toroidal world with kinematics
\[
\mathbf{x}_i(t{+}1)=\mathbf{x}_i(t)+v_i(t)[\cos \theta_i(t),\sin \theta_i(t)]\,\Delta t \ (\mathrm{mod}\ \Omega).
\]
Headings blend according to
$
\theta_i(t{+}1)=\arg\!\Big(\beta e^{\mathrm{j}\theta_i(t)}+(1{-}\beta)e^{\mathrm{j}\theta^\star_i}\Big),
$
where $\theta^\star_i=\theta(\mathbf{x}_i,t)$ and $\beta\in[0,1]$ controls inertia. Speeds follow either an exponential moving average (EMA) update $v_i(t{+}1)=\rho v_i(t)+(1{-}\rho)v_i^\star$ which smooths out abrupt changes, or an Ornstein-Uhlenbeck-like relaxation $v_i(t{+}1)=v_i(t)+\beta(v_i^\star-v_i(t))+\sigma\xi_t$ which adds small mean-reverting noise around the Perlin target. Because independent Perlin stacks drive headings and speeds, nearby agents share trends without collapsing into identical motion, delivering the high-coherence but diverse patterns observed in Figure~\ref{fig:behavior_fields}.

\begin{figure}[t]
    \centering
    \includegraphics[width=\linewidth]{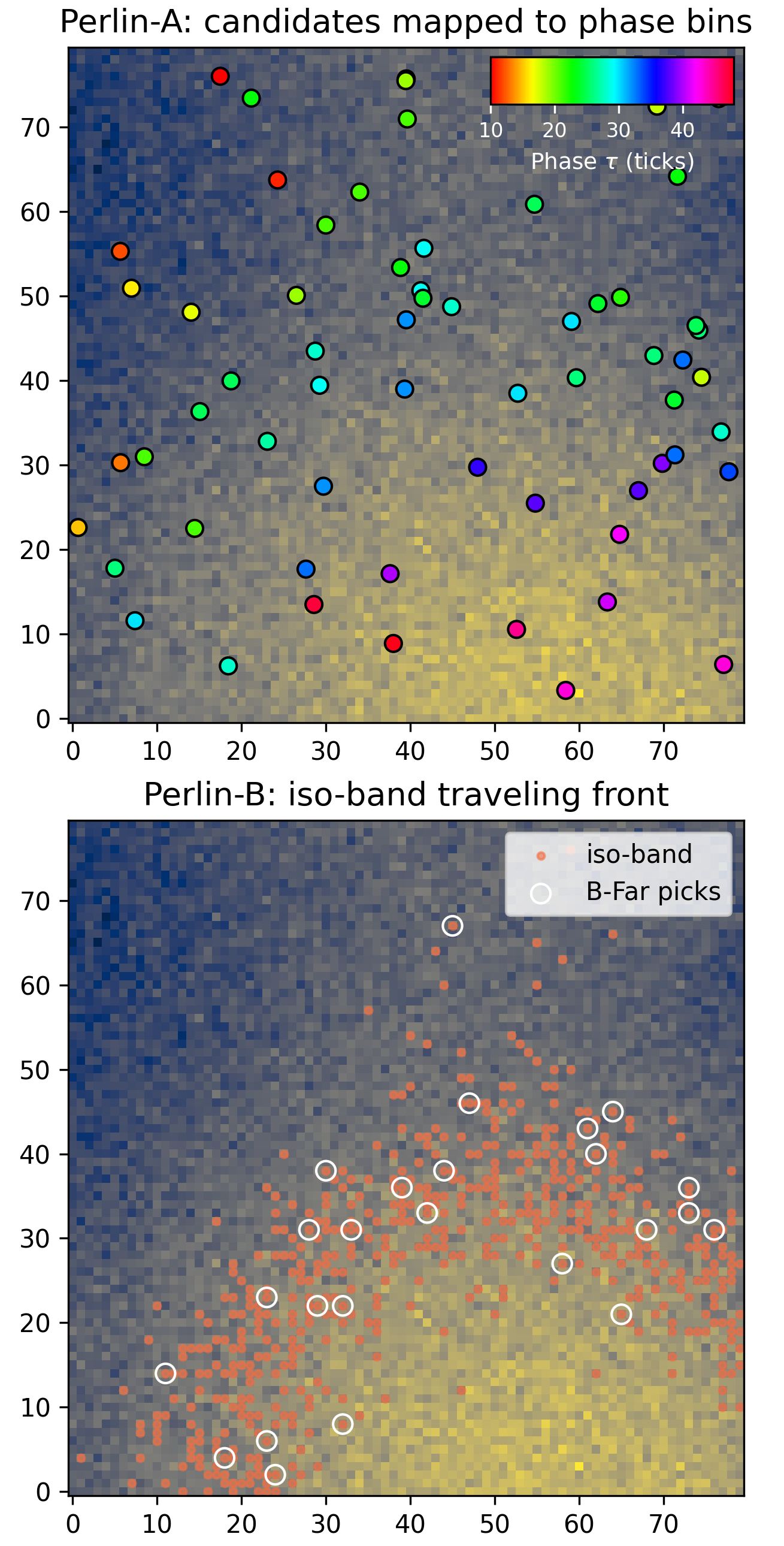}
    \caption{Spawn policies.     
    Top: Perlin-A (space$\rightarrow$time) samples candidate locations (dots), reads their Perlin values, and maps them to phase bins $\tau$ (dot colours); when the cycle time reaches $\tau$, those locations are proposed for spawns.
    Bottom: Perlin-B (time$\rightarrow$space) takes the current cycle time, selects an iso-band of positions with similar Perlin value (orange points), and picks well-separated B-Far sites (white circles), so a spawn front sweeps across the map as time advances.}
    \label{fig:spawn_policies}
\end{figure}

\subsection{Direction II: Activation Timing for Actions and Spawns}

\begin{figure*}[!t]
    \centering
    \includegraphics[width=\linewidth]{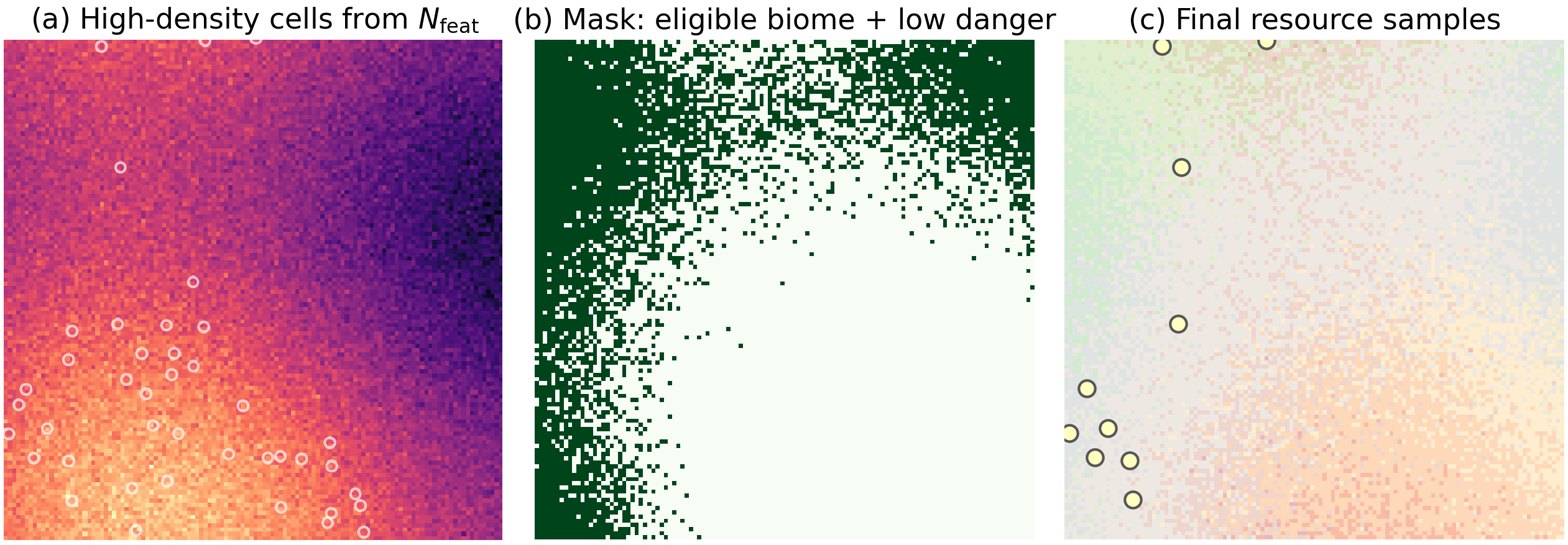}
    \caption{Single-feature sampling. Panel (a) shows the density gate derived from $N_{\mathrm{feat}}$. Panel (b) visualizes the discrete mask (eligible biome and low danger). Panel (c) overlays the surviving resource points, showing how one feature class is produced by intersecting the continuous gate with the discrete belts.}
    \label{fig:world_resource}
\end{figure*}

\begin{figure}[t]
    \centering
    \includegraphics[width=\linewidth]{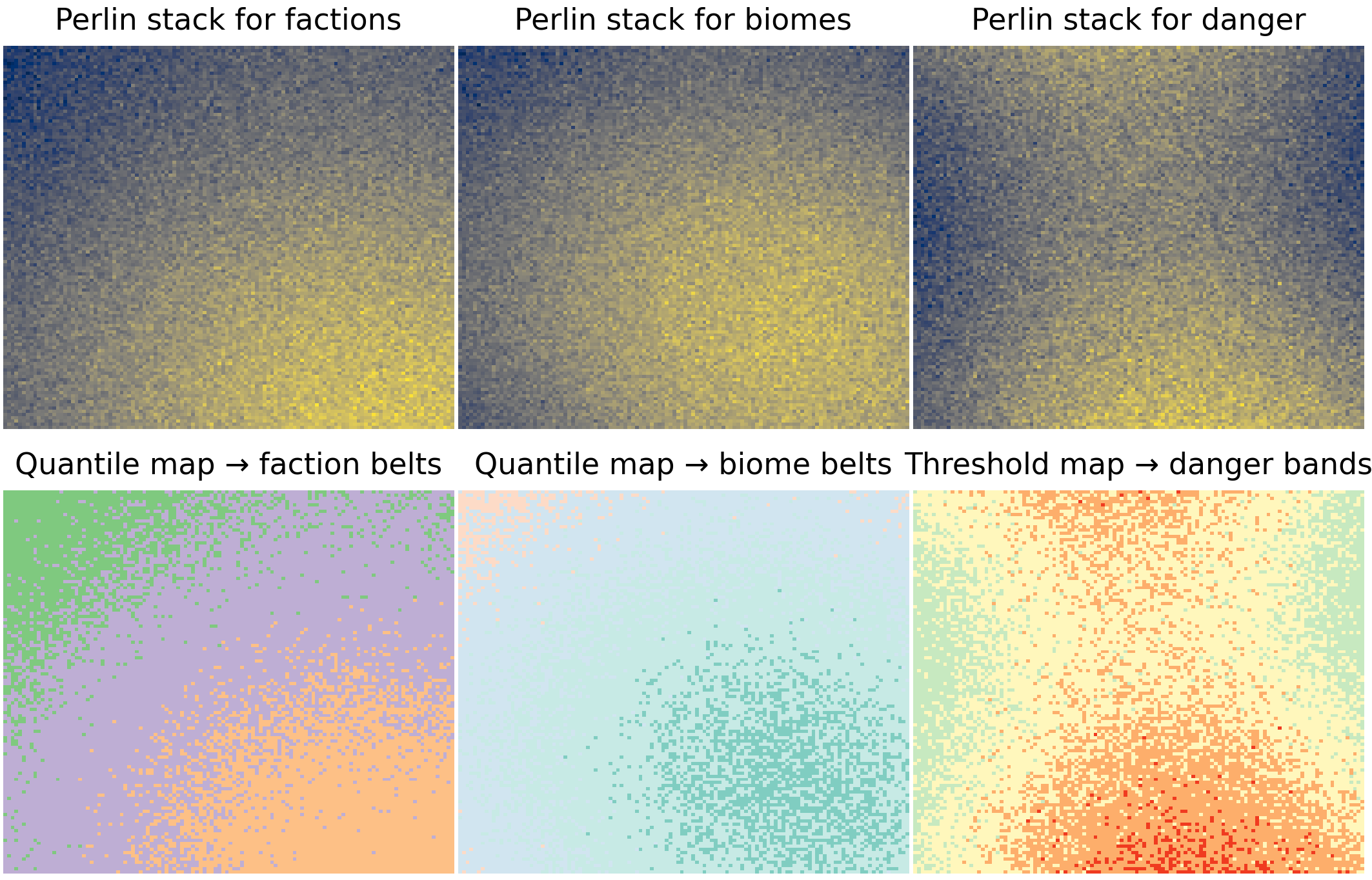}
    \caption{Discrete branch. Each categorical layer (factions, biomes, danger) samples an independent low-frequency Perlin stack (top row) and applies region-aware quantile or threshold mappings (bottom row), producing contiguous belts that satisfy design histograms.}
    \label{fig:world_discrete}
\end{figure}

Activation Timing modulates when and where behaviors start by sampling the same Perlin substrate at agent locations. While the underlying field is fully spatio-temporal $n(\mathbf{x}, t)$ over $\mathbf{x}\in\Omega\subset\mathbb{R}^2$, Figure~\ref{fig:activation_hazard} visualizes the mechanism on a 1D spatial slice for clarity (2D in practice). At each location we read a Perlin noise value, process and normalize it into either a hazard rate or a phase on a global cycle (e.g., a 60\,s loop), so that \textbf{an agent's position determines when in each cycle it is most likely to activate} and, symmetrically, \textbf{when and where the environment proposes new spawns}. A persistent field $n(\mathbf{x}, t)$ drives either hazards or phase-aligned activations. The hazard variant computes
\[
\lambda_i(t)=\lambda_0\big(\varepsilon+(1-\varepsilon)n(\mathbf{x}_i(t), t)\big),\qquad
p_i(t)=1-\exp(-\lambda_i(t)\Delta t),
\]
followed by global mean-normalization $\lambda_i\leftarrow \gamma \lambda_i$ with $\gamma$ chosen so that the average intensity matches $\lambda_{\mathrm{target}}=\lambda_0\big(\varepsilon+0.5(1-\varepsilon)\big)$. The phase variant samples $u_i=n(\mathbf{x}_i,t_0)$ at cycle start, converts it to a phase $\tau_i=\lfloor u_iT_{\mathrm{cycle}}\rfloor$, and assigns a circular Gaussian kernel $\kappa(\Delta)=\exp(-\tfrac{1}{2}(\Delta/\sigma)^2)$ around $\tau_i$. As visualized in Figure~\ref{fig:activation_hazard}, the final activation rate is a convex mixture
\[
\lambda_i^{\mathrm{hyb}}(t)=(1-\alpha)\lambda_i(t)+\alpha\,\lambda_0\big(\varepsilon+(1-\varepsilon)\kappa(\Delta\phi_i(t))\big),
\]
where $\Delta\phi_i$ measures the distance between the current cycle angle and $\tau_i$. This produces bright bands that travel across the spatial axis without collapsing into a single global wave, providing smooth starts, bounded burstiness, and predictable duty cycles.

For spawn placement under replenishment and cooldown constraints, we instantiate two complementary policies that couple space and time differently, as summarized in Figure~\ref{fig:spawn_policies}. \emph{Perlin-A} (space$\rightarrow$time) samples a static $n(\mathbf{x})\in[0,1]$, assigns $\tau_m=\lfloor n(\mathbf{x}_m)T_{\mathrm{cycle}}\rfloor$ to each candidate $\mathbf{x}_m$, and proposes all sites sharing the current phase (optionally thinned by $(1-\varepsilon)$). \emph{Perlin-B} (time$\rightarrow$space) maps the cycle angle $\theta$ to a level $\ell(\theta)$, forms an iso-band $S(\theta)=\{\mathbf{x}: |n(\mathbf{x})-\ell(\theta)|\le\varepsilon\}$, and then applies a farthest-point heuristic $\mathbf{x}^\star=\arg\max_{\mathbf{x}\in S(\theta)}\min_{\mathbf{q}\in E}\|\mathbf{x}-\mathbf{q}\|_2$ against existing entities $E$. Perlin-A is low-cost and strongly reproducible, whereas Perlin-B adapts spatial coverage via traveling iso-bands while maintaining smooth load because the band advances deterministically with $\theta$.

\subsection{Direction III: Type-Feature Generation for Synthetic Worlds}

Type-Feature Generation applies the dual-field paradigm to world layouts. Here, different Perlin fields are normalized and sliced into value ranges that map to discrete feature bins, so that each range of the field corresponds to a specific faction, biome, danger tier, or placement rule and directly controls where those types appear. Figure~\ref{fig:world_discrete} highlights how multiple Perlin noises are combined: three independent $N_{\mathrm{type}}$ stacks, each with its own octave schedule and seed offsets, are quantile-mapped within administrative regions to produce faction and biome belts, while ordered thresholds carve danger gradients. Formally,
$
\text{Map}_{\mathrm{disc}}(\mathbf{x})=\mathrm{bin}\big(Q_{\mathrm{region}}(N_{\mathrm{type}}(\mathbf{x}))\big)
$
ensures each administrative region receives the intended mix, while continuous attributes obey monotone transforms such as $N_{\mathrm{feat}}(\mathbf{x})\mapsto[\underline{h},\overline{h}]$ for HP multipliers. Height and auxiliary layers use their own seeds to sculpt ridges and coasts without disturbing discrete partitions.

Point placements arise from inhomogeneous Poisson processes constrained by the discrete masks and per-class quotas. Figure~\ref{fig:world_resource} focuses on the resource layer to keep the narrative concise: panel (a) highlights high-density cells gated by $N_{\mathrm{feat}}$; panel (b) applies the discrete mask (only specific biome bands at low danger tiers); panel (c) displays the surviving samples atop faction/danger backgrounds. Other feature classes, like wildlife, enemies, landmarks, follow analogous steps with their respective masks and quotas. Because every layer is tied to the same seed bundle $\{layout, noise, place\}$ with salted substreams (hashing the master seed with a per-layer tag to obtain independent deterministic RNG streams), rerendering with the same seed reproduces identical layouts and placements, enabling deterministic iteration while still benefiting from Perlin's spatial coherence. Layers are independent by default but may share seeds when correlation (e.g., biome–faction) is desired.

\section{Experiments}
\label{sec:experiments}

We evaluate the three parameter-field families introduced in Section~\ref{sec:method}. Unless stated otherwise, the spatial domain is a square torus of side $L=1000$ (meters), time is discrete with unit step $\Delta t=1$, and results are aggregated over 20 independent seeds.

\subsection{World model and assumptions}
In our experiment settings, agents are neutral NPCs evolving in a continuous 2D world under Perlin-driven parameter fields; players in our spawn study are scripted remover agents that eliminate entities within a kill radius. Experiments differ only in what is controlled: (i) \textbf{Behavior Parameterization} isolates motion (headings/speeds) for a fixed population; there are no admissions/removals and no players; (ii) \textbf{Activation Timing} has two substudies: \emph{(A) Perlin-driven Action Timing} triggers start-events (movement) on existing agents via hazard/phase (or hybrid) drivers with a fixed population, and \emph{(B) Perlin-based Spawn Placement} proposes spawn candidates via space-time Perlin policies in a non-toroidal world while a replenishment controller enforces per-cycle quotas and cooldowns to track target population under eliminations (player-driven or stochastic despawn); (iii) \textbf{Type-Feature Generation} is a content-generation study that produces discrete layouts (factions/biomes/danger) and continuous features (densities/rarities) and renders multi-view maps, rather than simulating agent dynamics.

\subsection{Behavior Parameterization}

We first evaluate how dual Perlin heading and speed fields modulate large neutral crowds to achieve locally coherent yet globally de-locked motion. Agents move on a continuous $2$D torus, integrating the velocity updates from Section~\ref{sec:method}. We run three crowd scales (from a few hundred to a few thousand agents, over horizons corresponding to 6-18 minutes of simulated time) and replicate each configuration over 20 independent seeds for paired comparisons.

Our default Perlin configuration uses two independent multi-octave fields with slow temporal drift: one field is mapped to heading targets, the other to speed targets. At each time-step, headings are updated by vector-averaging previous and target directions to avoid wrap-around artifacts, and speeds are exponentially smoothed to provide inertia while preserving local coherence. We additionally sweep Perlin spatial and temporal scales (frequency, octave count, persistence, drift) and population size to test robustness across crowd densities and correlation lengths.

We compare against six representative baselines under matched seeds, horizons, and logging: \emph{Perlin-single} (a single scalar field drives both heading and speed), \emph{URW} (uncorrelated random walk with fixed speed), \emph{OU-heading} \cite{uhlenbeck1930theory} (Ornstein-Uhlenbeck process in heading space), \emph{Curl-noise} (divergence-free flow from a Perlin potential), \emph{Vicsek} \cite{vicsek1995novel} (neighbor alignment within a fixed interaction radius), and \emph{Piecewise} (coarse piecewise-constant vector grid with interpolation). Baseline parameters are tuned to roughly match average speed and avoid trivial failures. Full evaluation details in Appendix~\ref{app:exp-details}.

\subsection{Activation Timing}

We next evaluate temporal fields that modulate when and where activity concentrates under global rate budgets, quotas, and cooldowns. We consider two complementary experiments: (A) per-agent action starts driven by Perlin hazard/phase (or hybrid) fields on a toroidal world with fixed agent populations, and (B) spawn placement driven by space-time Perlin couplings in a non-toroidal world where a replenishment controller tracks target populations under eliminations. All configurations are run at multiple spatial and population scales with 20 fixed seeds per setting. Detailed numeric parameters and hyperparameter grids are listed in Appendix~\ref{app:exp-details}.

\subsubsection{Perlin-driven Action Timing}

For per-agent actions, we test whether Perlin-based hazard and phase drivers produce locally synchronized yet globally de-locked start events relative to stochastic, deterministic, and constraint-oriented baselines. A persistent Perlin field with slow drift is sampled at agent positions and mapped either to local hazard rates or to cycle phases; an optional convex mixture between the two controls the sharpness and contrast of traveling fronts over each cycle. We run three agent scales (hundreds to many thousands of agents on maps of increasing physical extent) and sweep Perlin hyperparameters around a nominal configuration.

Baselines cover a range of scheduling styles: constant-rate \emph{Poisson} triggers, \emph{Filtered random} proposals with local suppression, \emph{Fixed} periodic schedules with jitter, two \emph{Constraint}-based schedulers (token and round-robin variants), a global \emph{Sinusoid} modulation of the rate, and a \emph{Hawkes-inhibitory} point process \cite{hawkes1971spectra} with local self-inhibition. Each baseline is tuned to match the overall mean rate of the Perlin driver. We quantify timing regularity via inter-event-interval statistics (mean, variance, coefficient of variation, duty cycle, gap percentiles) and burstiness indices (including Fano factor \cite{fano1947ionization}), temporal smoothness via second-difference energy and spectral high/low-frequency ratios of the active-count series, spatial balance via regional variation and Moran’s~$I$ \cite{moran1950notes}, and space-time patterning via front coherence and second-order point-process summaries (Ripley’s $K(r)$ and pair-correlation $g(r)$) \cite{ripley1976second,baddeley2016spatial}.

\subsubsection{Perlin-based Spawn Placement}

For spawn placement, we assess two Perlin space-time couplings under replenishment quotas, cooldowns, eliminations, and player interactions. In \emph{Perlin-A} (space$\to$time), a static Perlin field assigns a cycle phase to a stratified set of candidate sites, producing phase-binned micro-batches each cycle. In \emph{Perlin-B} (time$\to$space), the cycle angle selects an iso-band of the Perlin field that sweeps across the map; proposals are drawn from this band, either uniformly or via farthest-point sampling (B-Far) to adapt to existing entities and reduce local congestion. We simulate square non-toroidal worlds at three scales (increasing map bounds, target populations, and per-cycle quotas) while holding monster and player dynamics fixed.

We benchmark Perlin-A and Perlin-B against \emph{Uniform random} placement, \emph{Filtered random} with safety and spacing filters, \emph{Poisson-disk} (blue-noise) placements, \emph{MVN+Poisson} (mixtures of Gaussians with Poisson counts), heuristic \emph{Facility-location} scoring, and a purely temporal \emph{Sinusoid} with no spatial structure, all rate-matched to the same target populations and quotas. Metrics emphasize spatial coverage and balance (coverage distance, regional variation, spatial autocorrelation, nearest-neighbor and pair-correlation summaries), temporal stability (inter-spawn variability, spectral smoothness, front coherence), and efficiency (per-tick runtime, spawns and eliminations per cycle, load variance). Exact thresholds, window sizes, and coverage-sampling schemes are detailed in Appendix~\ref{app:exp-details}.

\subsection{Type-Feature Generation for Synthetic Worlds}

Finally, we instantiate Type-Feature Generation as a New World-style synthetic territory builder that decouples discrete layouts (factions, biomes, danger bands, content types) from continuous attributes (density multipliers, rarity, power scaling). The world is a square island sampled on a regular grid, and all layers are driven by a seed bundle that deterministically maps a user-visible integer seed to three substreams $(layout, noise, place)$, ensuring run-level reproducibility and per-layer independence. See Appendix~\ref{app:newworld} for the lore-aware world prior and presets that parameterize our generator.

We employ three territory templates inspired by different regions (\emph{WindswardLike}, \emph{ShatteredMountainLike}, \emph{EdengroveLike}) that specify high-level design priors: per-region histograms for faction/biome/danger mixes, quotas and exclusion radii for resources, wildlife, enemy families, and landmarks, and feature ranges per danger band (e.g., hp/dps scaling, elite/boss likelihoods). Low-frequency $N_{\texttt{type}}$ fields drive discrete maps via regional quantile mapping to hit the specified mixes and maintain contiguity, while mid/high-frequency $N_{\texttt{feat}}$ fields drive continuous intensities via monotone transforms. Placements for each content class are realized by Poisson processes gated by these intensities and masks, with per-class minimum distances and quotas enforcing coverage and rarity.

Each run produces a fixed five-view bundle of rendered maps: an \emph{overview} (factions, danger outlines, landmarks, rare content), \emph{resources-by-biome}, \emph{enemies-by-danger}, \emph{landmarks-by-faction}, and zoomed quadrants that highlight resource peaks, enemy peaks, and landmark clusters. Optional vector exports carry per-point attributes for downstream analysis and integration. In our experiments we showcase three seeds (one per template) using a common visualization layout so that differences reflect only template priors and random seeds, not rendering choices. The exact resolutions, noise-layer specifications, danger thresholds, and symbol-density settings are documented in Appendix~\ref{app:exp-details}.

\section{Results}
\label{sec:results}

We evaluate three application directions: Behavior Parameterization for crowd motion, Activation Timing for per-agent actions and spawns, and Type-Feature Generation for synthetic worlds. Throughout, we emphasize the most important empirical findings in the main text and refer the reader to Appendix~\ref{app:extended-results} for full metric grids, additional diagnostics, and extended discussion.

\begin{figure}[t]
  \centering
  \includegraphics[width=\columnwidth]{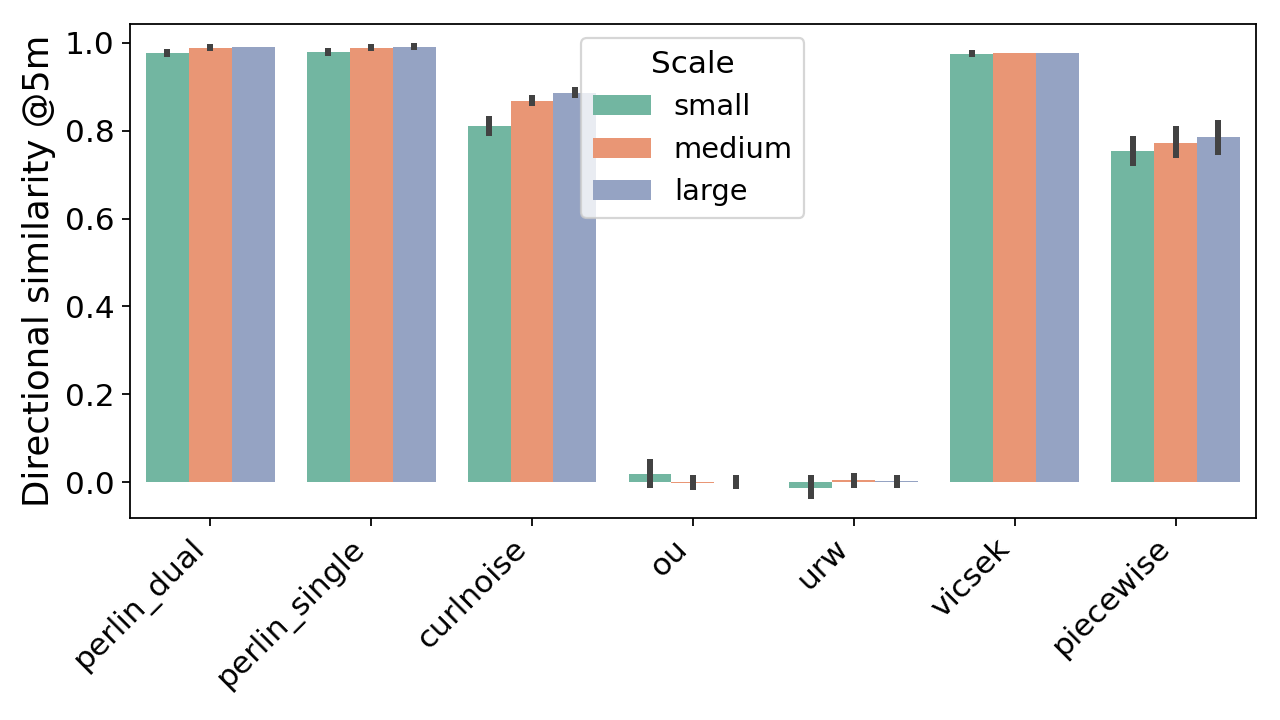}
  \vspace{-0.7em}
  \caption{5.1 Directional similarity at 5\,m \(S_{\mathrm{dir}}@5\uparrow\) across scales. Perlin variants are consistently highest while remaining lightweight.}
  \label{fig:bp_sdir}
\end{figure}

\begin{figure}[t]
  \centering
  \includegraphics[width=\columnwidth]{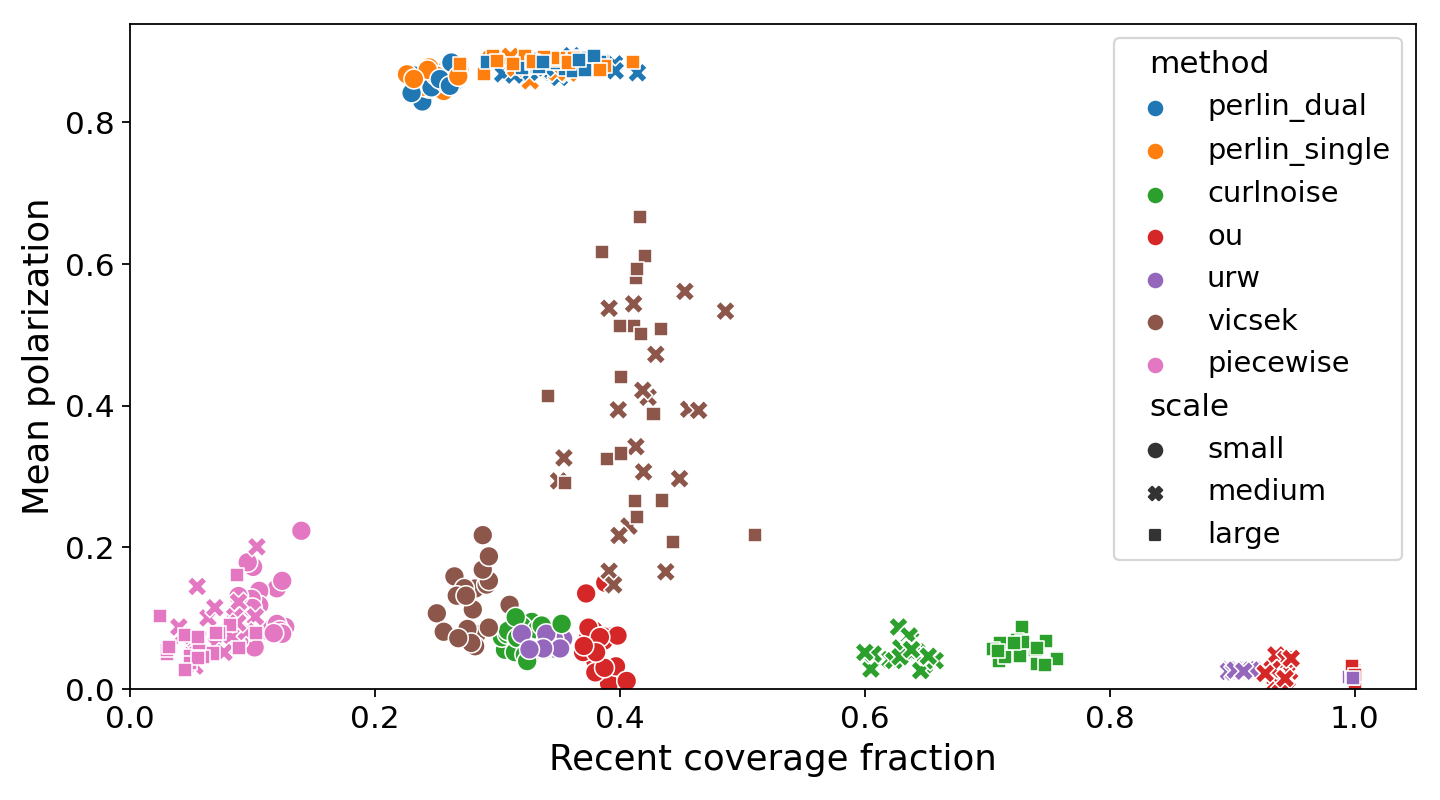}
  \vspace{-0.7em}
  \caption{5.1 Coverage vs.\ polarization (marker: seed$\times$scale run).
  Perlin emphasizes coherence with moderate coverage; stochastic baselines explore more but lose local structure.}
  \label{fig:bp_tradeoff}
\end{figure}

\subsection{Behavior Parameterization}

Across all three crowd scales, dual Perlin fields produce the strongest local spatial coherence while maintaining smooth motion at lightweight runtime. Figure~\ref{fig:bp_sdir} reports directional similarity at 5\,m, \(S_{\mathrm{dir}}@5\), showing that both \texttt{perlin\_dual} and \texttt{perlin\_single} achieve values near 1.0 at all scales (e.g., almost perfectly aligned headings for nearby agents, which produces lane-like local flow and avoids visually noisy counter-rotations), clearly exceeding \texttt{curlnoise} and far above stochastic baselines such as \texttt{ou} and \texttt{urw}, whose \(S_{\mathrm{dir}}@5\) values are close to zero. \texttt{vicsek} also attains high local similarity, but requires explicit neighbor alignment and is orders of magnitude more expensive to run. Together with low jerk statistics (reported in Appendix~\ref{app:extended-results-behavior}), these results confirm that Perlin fields naturally induce locally coherent, temporally smooth flows without explicit communication between agents.

The main trade-off is between exploration and global alignment. Figure~\ref{fig:bp_tradeoff} plots recent coverage fraction against polarization for each seed$\times$scale run. Perlin methods occupy a distinctive regime of high coherence and moderate coverage with relatively high polarization: they cover substantially more area than \texttt{piecewise} fields while preserving structure, but concede maximum coverage to \texttt{ou} and \texttt{urw}, which behave like nearly structureless diffusion with near-zero local similarity. This matches the design intent of Behavior Parameterization: we prioritize visually coherent flows and smooth trajectories over purely maximizing visited area. Runtime measurements (Appendix~\ref{app:extended-results-behavior}) show that Perlin updates remain in the same order as lightweight baselines and far below neighbor-based \texttt{vicsek}, and that both local coherence and jerk remain stable from small (200 agents) to large (3200 agents) crowds. The main limitation is elevated global polarization due to field-aligned motion; Section~\ref{sec:method} and Appendix~\ref{app:extended-results-behavior} discuss simple mitigations such as mixing multiple orientation fields, adding curl components, or injecting low-variance jitter to de-lock headings.

\begin{figure}[t]
  \centering
  \includegraphics[width=\columnwidth]{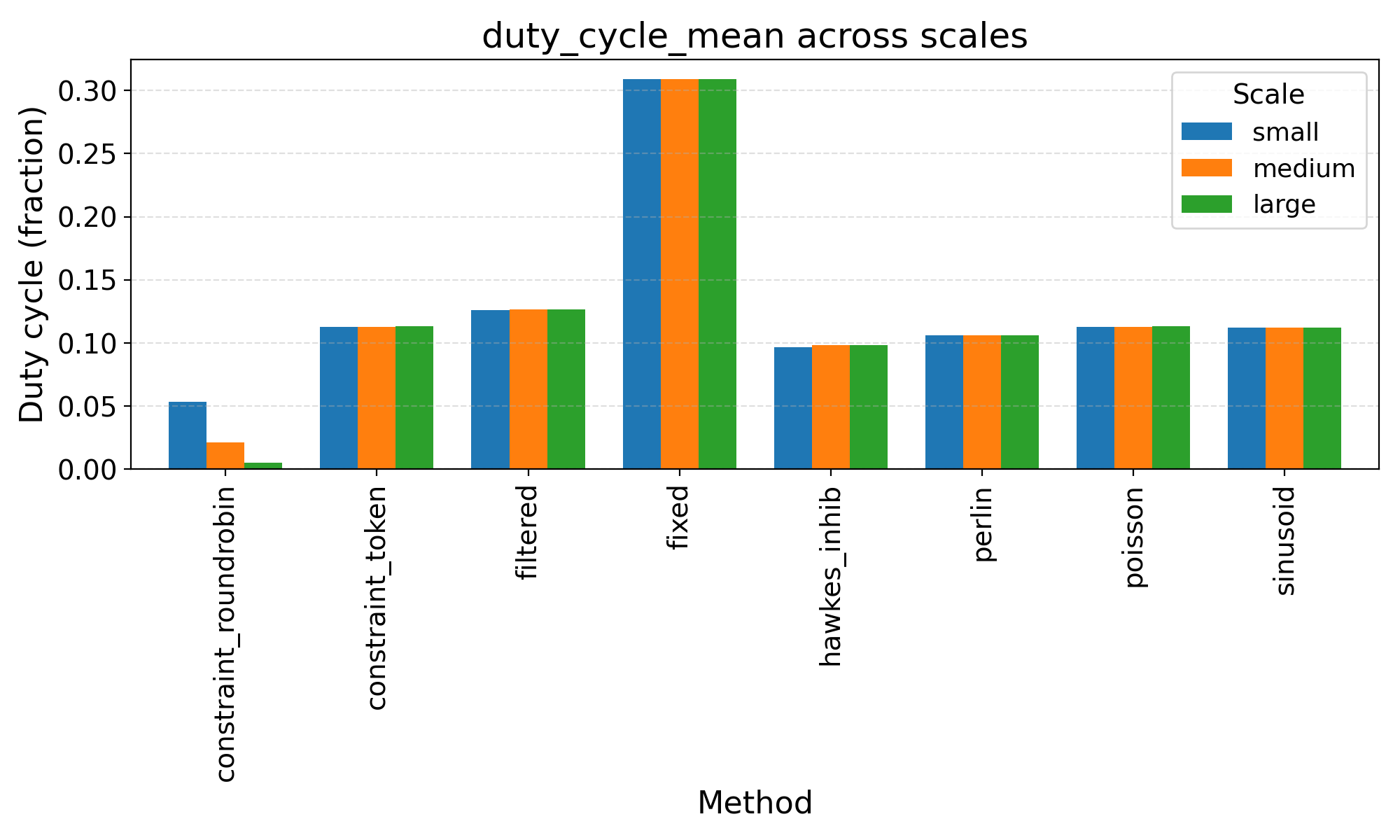}
  \vspace{-0.5em}
  \caption{5.2.1 Duty cycle (mean fraction of active agents) across scales. Perlin matches the target rate similarly to Poisson/Sinusoid/Filtered; token/roundrobin reflect capacity limits; fixed is not rate-matched and therefore runs at a higher duty.}
  \label{fig:at_duty_cycle}
\end{figure}

\begin{figure}[t]
  \centering
  \includegraphics[width=\columnwidth]{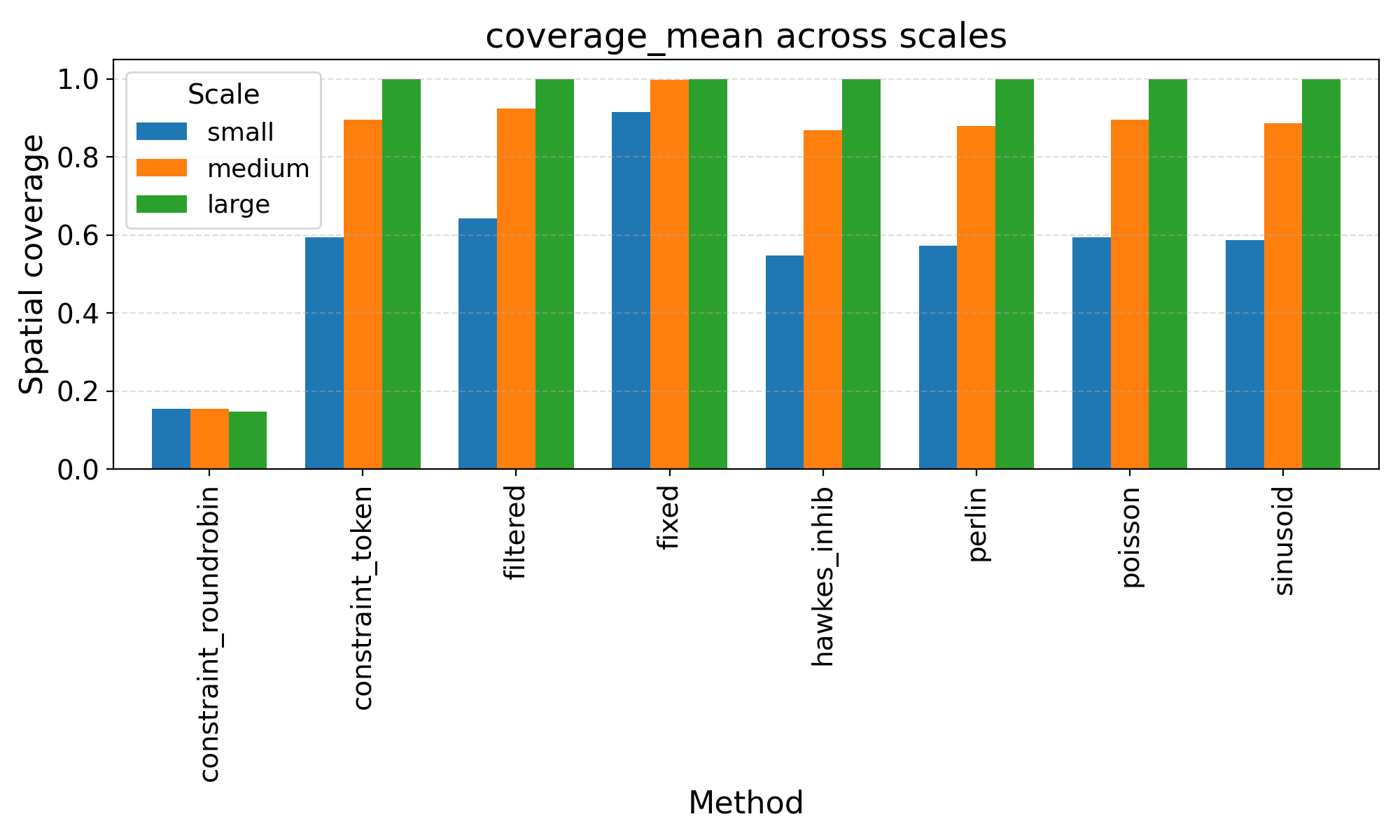}
  \vspace{-0.5em}
  \caption{5.2.1 Spatial coverage across scales. Perlin scales up cleanly and approaches saturation at Large, comparable to stochastic baselines.}
  \label{fig:at_coverage}
\end{figure}

\subsection{Activation Timing}

\subsubsection{Perlin-driven Action Timing}

\begin{figure}[t]
  \centering
  \includegraphics[width=\linewidth]{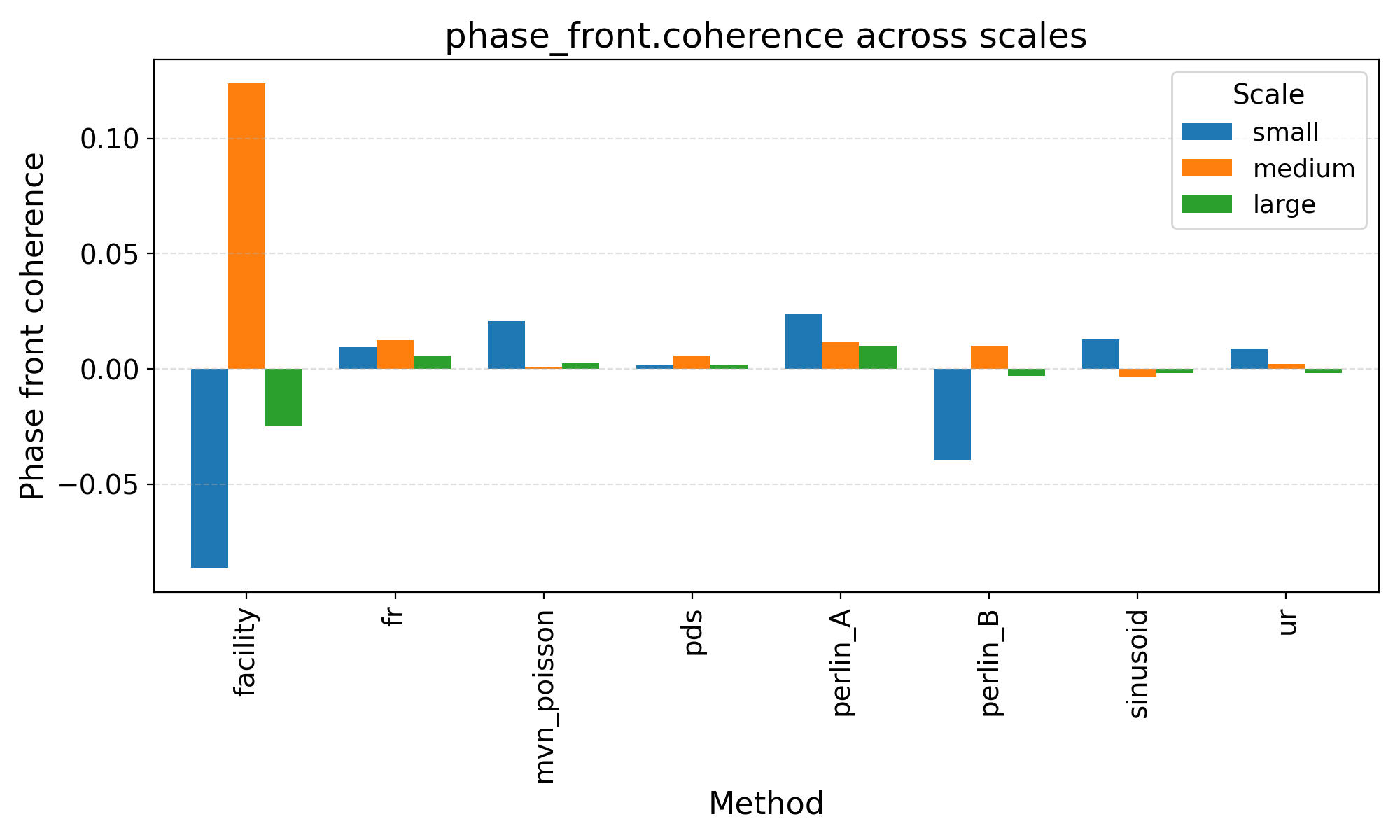}
  \vspace{-6pt}
  \caption{5.2.2 Phase-front coherence across scales.
  Perlin-A remains positive and stable across scales; Perlin-B hovers near zero; some baselines flip sign (e.g., Facility).}
  \label{fig:spawn-coherence}
  \vspace{-4pt}
\end{figure}

\begin{figure}[t]
  \centering
  \includegraphics[width=\linewidth]{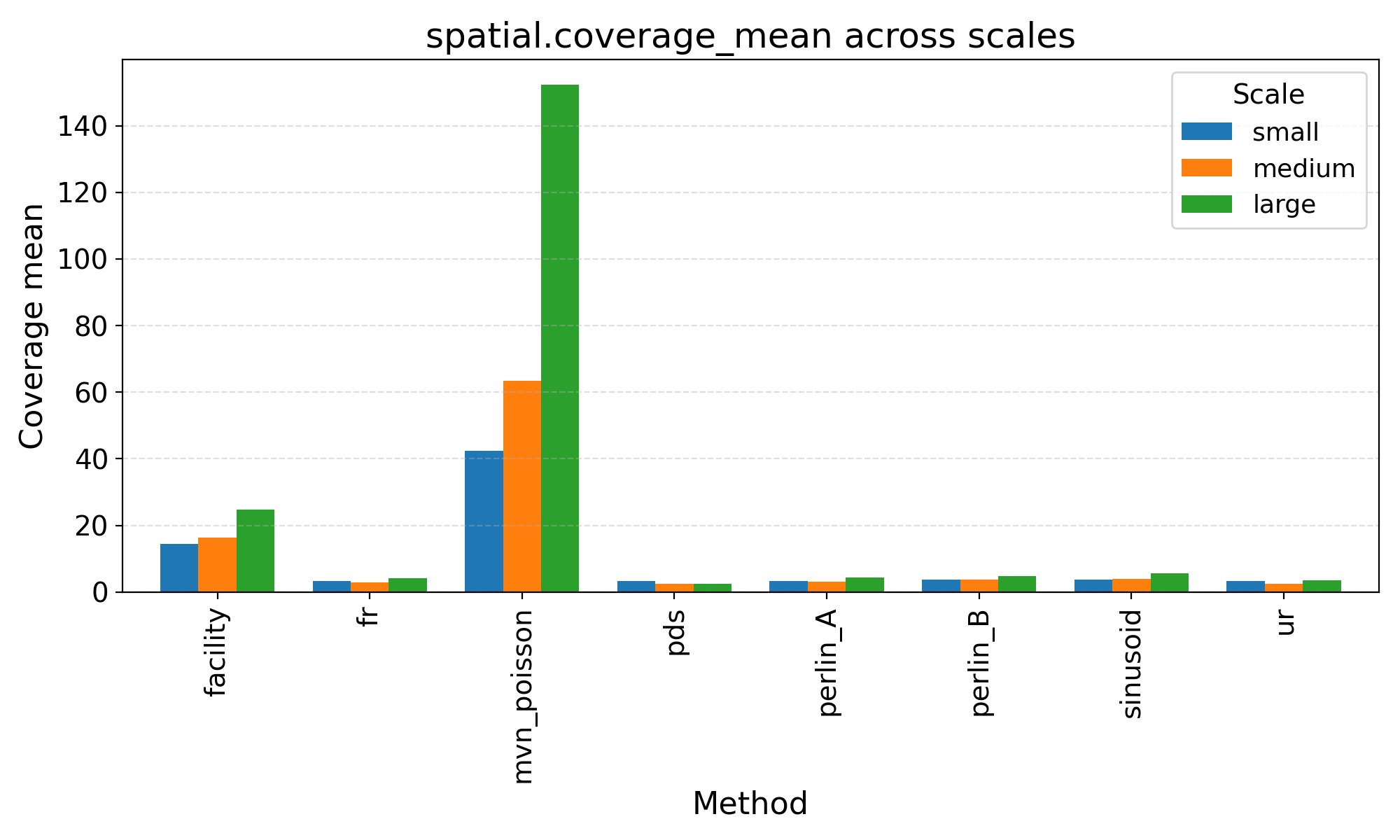}
  \vspace{-6pt}
  \caption{5.2.2 Coverage distance (mean; world units) across scales.
  Perlin-A/B achieve coverage distances in the same range as lightweight baselines (UR/PDS/FR). MVN+Poisson attains very large values due to its clustered density objective (not directly comparable to balanced-coverage goals).}
  \label{fig:spawn-coverage}
  \vspace{-6pt}
\end{figure}

Perlin hazard/phase scheduling achieves precise global rate control while producing locally coherent yet globally de-locked action starts. Figure~\ref{fig:at_duty_cycle} shows duty cycle across scales for all methods: among the rate-matched group (Perlin, Poisson, Filtered, Sinusoid, Hawkes-inhibitory, Constraint-Token), Perlin’s mean duty closely tracks the target and is very close to Poisson and Sinusoid, whereas Fixed runs at a much higher duty and Constraint-Roundrobin underutilizes capacity by design. This indicates that mean-rate normalization of the Perlin hazard field succeeds in enforcing the desired activation budget.

Temporal smoothness and burstiness are summarized in Appendix~\ref{app:extended-results-activation-action} via high/low-frequency energy ratios and Fano factors. We seek smooth, desynchronized action timing under fixed duty-cycle budgets, so we penalize high-frequency chattering and rigid global waves and treat Fano as a lightweight proxy for natural variability. Qualitatively, Perlin suppresses high-frequency chattering in the active-count series and maintains near-Poisson variability while avoiding the rigid global waves induced by Sinusoid and the capacity-driven sawtooths of Constraint-Roundrobin. At the same time, spatial coverage (Figure~\ref{fig:at_coverage}) increases monotonically with scale for all methods; Perlin’s coverage is competitive with Poisson and Filtered at Medium and saturates near unity at Large, indicating that localized temporal coherence does not impair long-run exploration under realistic agent densities. Runtime scales linearly with population and is higher than simple Poisson/Filtered baselines but comparable to more structured schedulers such as Hawkes-inhibitory and capacity-based variants (Appendix~\ref{app:extended-results-activation-action}). Overall, Perlin-based action timing offers a favorable balance between controllable duty cycles, smooth but unsynchronized temporal structure, strong spatial coverage, and moderate computational overhead.

\subsubsection{Perlin-based Spawn Placement}

\begin{figure*}[!t]
  \centering
  \includegraphics[width=.48\linewidth]{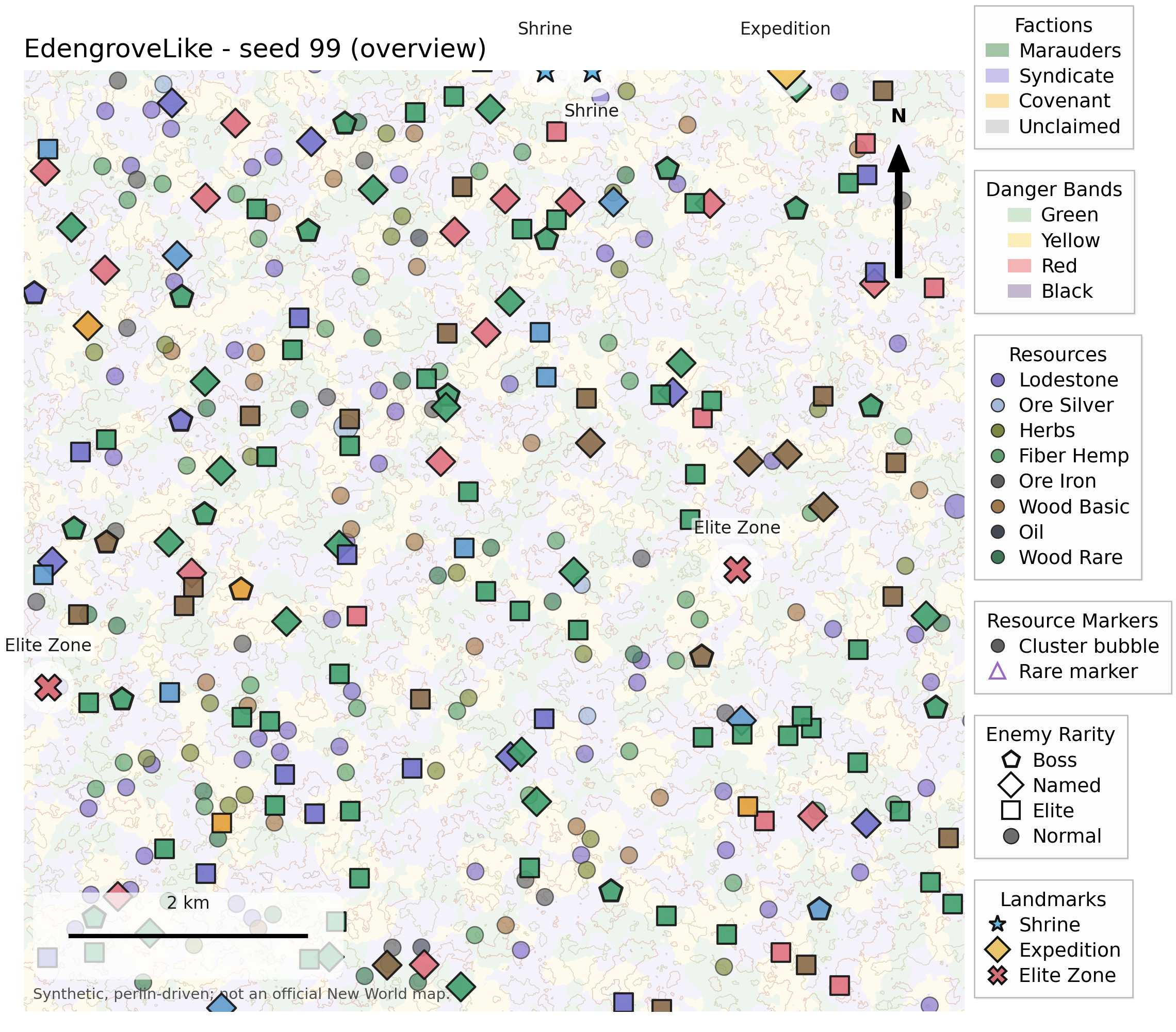}\hfill
  \includegraphics[width=.48\linewidth]{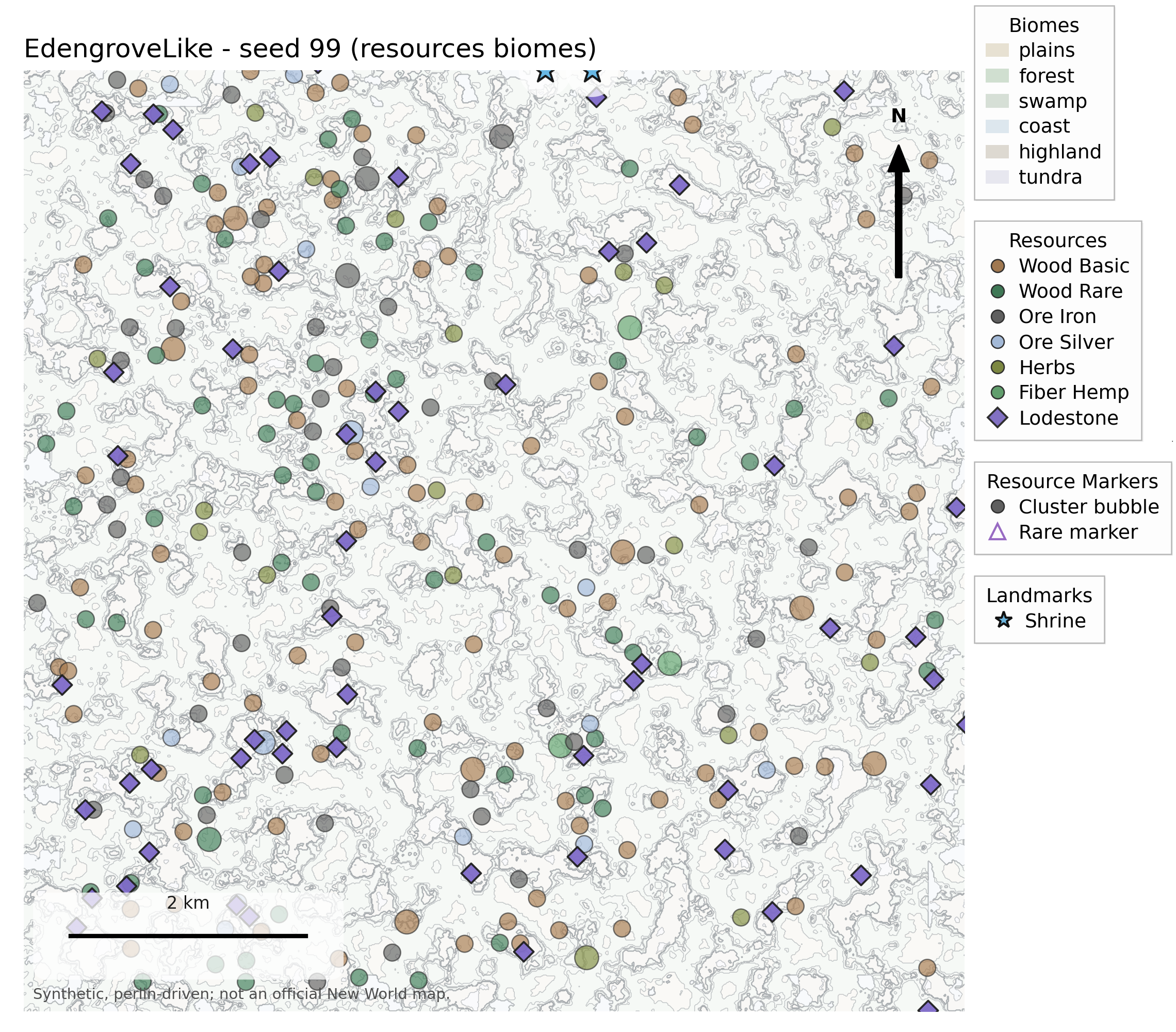}\\[4pt]
  \includegraphics[width=.48\linewidth]{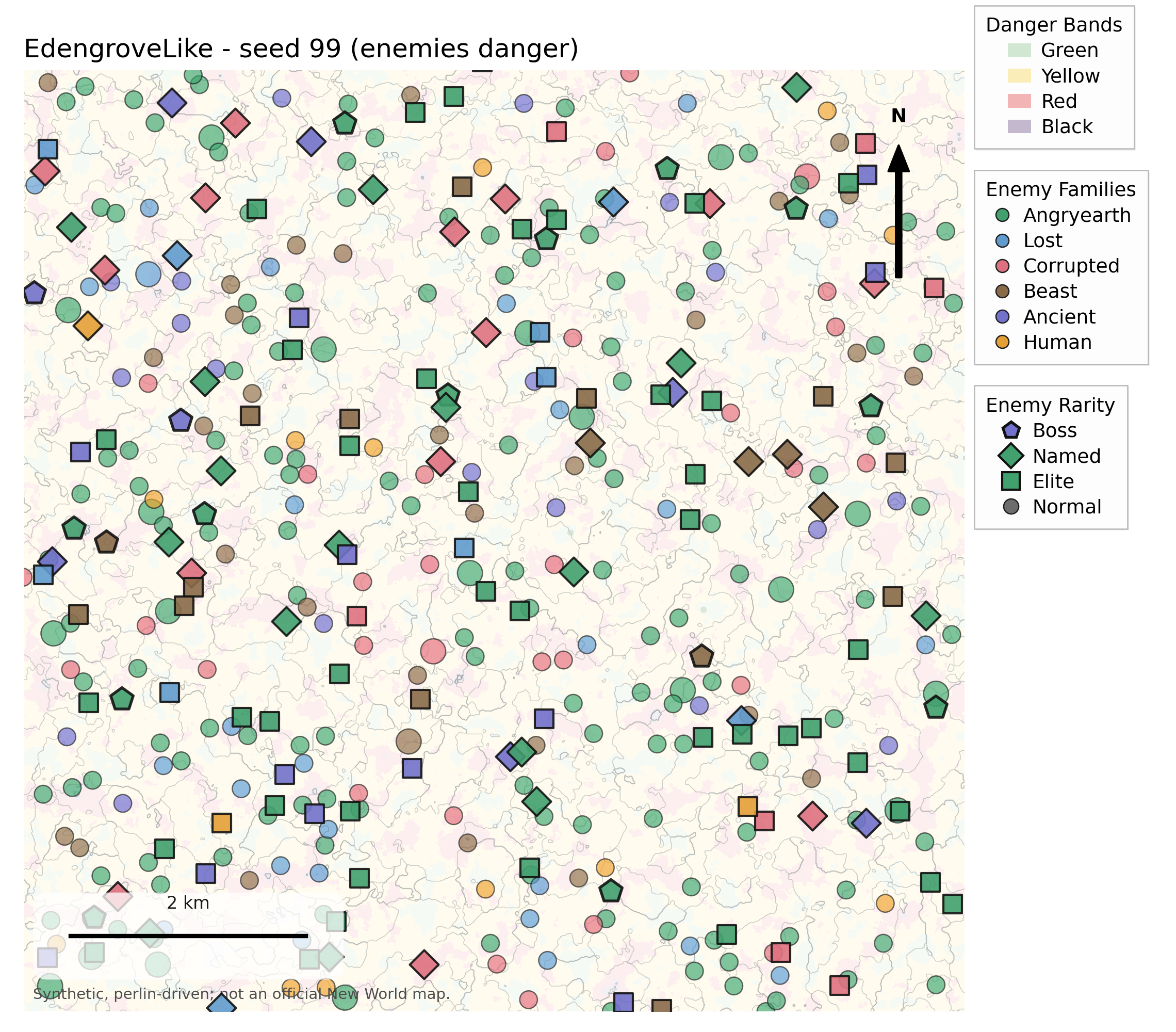}\hfill
  \includegraphics[width=.48\linewidth]{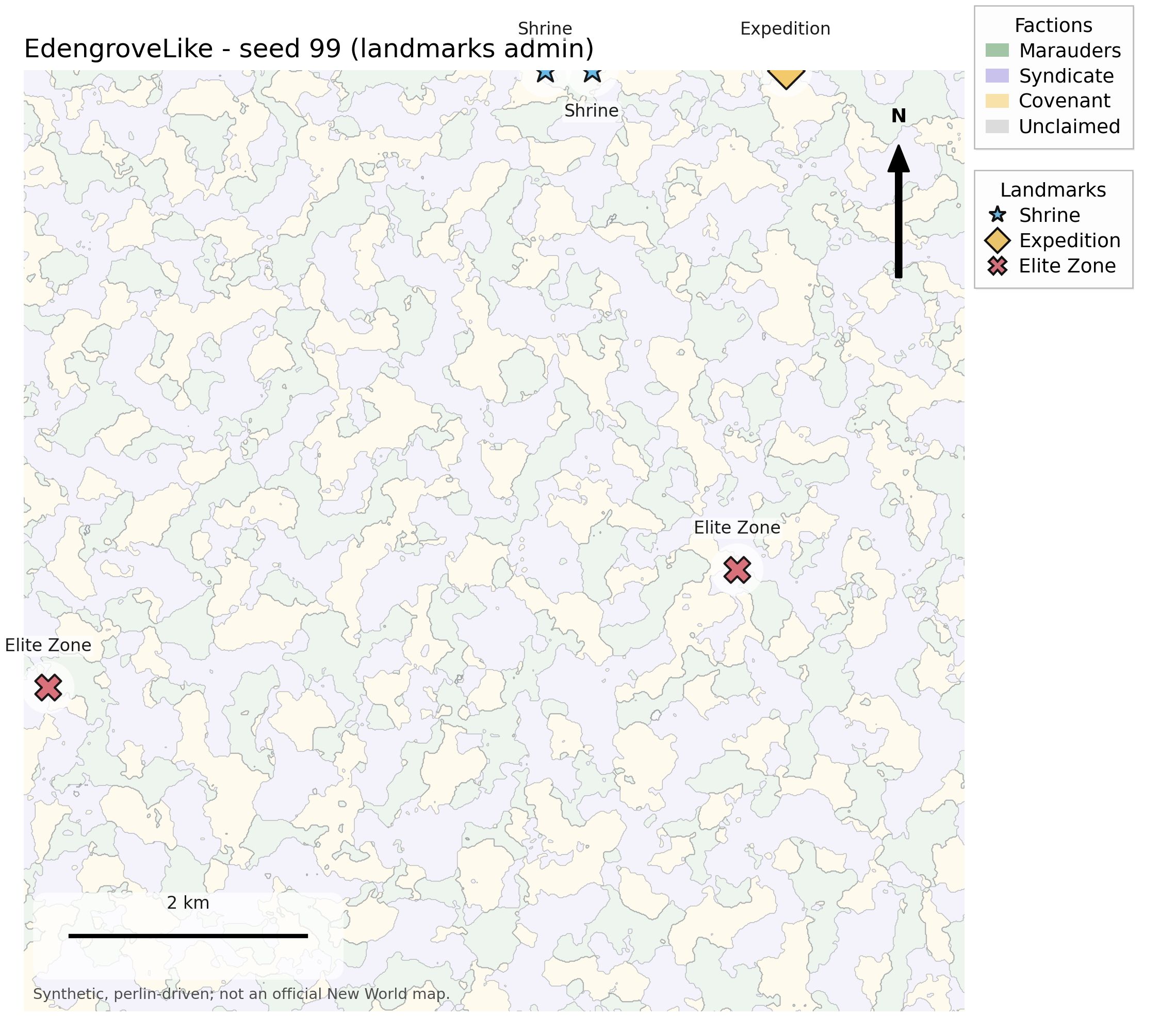}
  \caption{\textbf{Five-view bundle (four plates shown) for \texttt{EdengroveLike}, seed 99.}
  \emph{Top-left:} overview—danger bands and factions overlaid with resources, enemy rarities, and landmarks.
  \emph{Top-right:} resources over biomes.
  \emph{Bottom-left:} enemy families/rarities over danger bands.
  \emph{Bottom-right:} landmarks over faction control.
  Synthetic, Perlin-driven; not an official New World map.}
  \label{fig:nw_bundle}
\end{figure*}

For spawn placement, Perlin space-time couplings provide reliable fronts and coverage with smooth load and mid-pack runtime. Perlin-A (space$\to$time) assigns cycle phases to candidate locations, creating low-amplitude micro-batches that track a moving front. Figure~\ref{fig:spawn-coherence} shows that its phase-front coherence is small but positive and remains stable as we move from small to large maps, in contrast to methods like Facility whose coherence can flip sign as spatial patterns change. Perlin-B (time$\to$space) sweeps an iso-band across the Perlin field and focuses on balanced spatial exposure: its coherence is near zero, by design, and its coverage distances remain in the same range as lightweight baselines such as Uniform Random, Filtered Random, and Poisson-disk sampling (Figure~\ref{fig:spawn-coverage}), while the iso-band front and B-Far selection help avoid large uncovered gaps at larger scales.

Extended diagnostics in Appendix~\ref{app:extended-results-activation-spawn} show that Perlin-A and Perlin-B produce smoother inter-spawn intervals and lower load variance than cluster-oriented methods, while avoiding the high-amplitude bursts of Poisson-disk scheduling and the globally synchronized oscillations of Sinusoid. Runtime for both Perlin policies grows nearly linearly with problem size and remains in the middle of the pack: more expensive than the cheapest density-only baselines, but cheaper than heavily filtered or facility-location-based strategies at large scale. In practice, Perlin-A is best viewed as a lightweight way to obtain reproducible, gently moving fronts in space-time, whereas Perlin-B emphasizes balanced spatial coverage under stable load; both can be combined with downstream replenishment controllers and capacity constraints without modification.

\subsection{Type-Feature Generation}

Our Perlin-driven generator produces New World-style territories whose discrete layouts and continuous attributes are visually coherent and gameplay-plausible. Figure~\ref{fig:nw_bundle} illustrates the multi-view bundle for the \texttt{EdengroveLike} template (four plates shown). In the overview and landmarks plates, faction fills form contiguous territories with gently curving borders, danger bands appear as interpretable islands and corridors rather than salt-and-pepper speckles, and biome fills occupy broad belts and pockets.

The resources-by-biomes plate (top right) shows that mid/high-frequency feature fields gate densities in plausible ways: common resources (e.g., wood, hemp, herbs) follow biome expectations and appear in clusters rather than uniform scatter, while rare nodes concentrate in a handful of high-intensity pockets such as ridges or remote clearings. In the enemies-by-danger plate (bottom left), enemy families align with their intended habitats (e.g., Angry Earth in forest belts, Ancients in highlands/ruins), and rarity markers (elite/named/boss) track the underlying danger bands, with high-tier enemies mostly confined to Red/Black zones and only occasional elites seeding Green/Yellow zones. Landmarks (shrines, towns, expeditions, elite zones) occupy positions that are both readable on the map and navigationally meaningful: shrines cluster near junctions and corridors, expedition entrances sit in remote highland pockets, and elite zones lie in high-danger corridors with surrounding maneuvering space. Across plates, symbol density and legend layout are held fixed, and seed bundles map user-visible seeds to $(layout,noise,place)$ substreams, ensuring that reruns with the same seed reproduce both region layouts and point placements.

A detailed audit, including zoomed quadrants, quantitative quota errors, spacing statistics, and co-location rates, is provided in Appendix~\ref{app:extended-results-world}. We note the main limitations: some very sparse danger bands still reveal regularity where minimum-distance constraints dominate, the current height layer is generic and does not yet drive specific POI geometry or road/water networks, and ecology is limited to static priors rather than dynamic interactions such as predator-prey cycles. Nonetheless, the qualitative evidence supports our design: low-frequency type fields create quota-matching contiguous regions; mid/high-frequency feature fields create textured densities; and inhomogeneous Poisson placement with spacing rules turns these fields into readable, reproducible worlds.

\section{Conclusion}

In this paper, we introduced a noise field-driven framework for large-scale game AI coordination. We instantiated the framework in three directions. Quantitative experiments showed the intended trade-offs: high local coherence and temporal smoothness, competitive spatial/temporal coverage, and predictable budgets relative to baselines; while qualitative world results showed coherent layering and plausible content distributions. These findings validate that a noise field-driven method can balance efficiency with expressivity and offer a practical path to scalable, de-locked coordination that integrates cleanly with existing navigation and authoring pipelines.

Despite these strengths, several limitations remain. Firstly, our behavior parameterization and activation timing tracks focus on foundational capabilities (kinematic motion and spawning); many domain-specific behaviors within each track remain unexplored. Secondly, we evaluate large-scale control/motion/generation under simplified interaction assumptions; real productions often require additional constraints (e.g., collisions, reciprocal avoidance, designer overrides). Incorporating richer interaction without sacrificing efficiency is an open problem. Thirdly, we instantiate Perlin fields in 2D; while sufficient for our baseline needs, higher-dimensional noise (3D/4D or manifold-aware fields) offers substantial headroom. Fourthly, our assessments rely on quantitative metrics and qualitative inspection; ultimately, smoothness, believability, and world/playability are experiential and benefit from designer/player studies, especially since simple, efficient methods may suffice in practice depending on style and constraints.

We anticipate several future works aligned with these limitations. Within the same framework, we plan to extend behavior parameterization to drive NPC alertness, state/mood, and group tactics, and use activation timing for large-formation/army-level phase control. We will also combine noise fields with potential/flow fields and lightweight kernel filters to extract local summaries (e.g., density, gradient/curvature cues) that enable limited local communication while preserving frame predictability. Besides, we will explore higher-dimensional fields: volumetric/stacked 3D for multi-floor/voxel spaces and 4D (space-time) for diurnal/seasonal modulation, enabling smoothly evolving encounter/ecology schedules. Finally, we will develop evaluation protocols that better quantify aesthetics and playability (e.g., quota error, dispersion/fairness KPIs, perceived naturalness) and conduct designer/player user studies to validate experiential quality under matched frame budgets.

In sum, by leveraging the seedability, frequency control, and coherence of Perlin noise, we extend its traditional use in terrain and texture synthesis into a general representation for real-time, large-scale AI coordination and world structuring. We hope this framework serves as a foundation for future work that bridges procedural layout and runtime scheduling, integrates richer interactions, and advances controllable and believable game worlds.

\clearpage

\bibliographystyle{ACM-Reference-Format}
\bibliography{references}

\appendix
\section{New World-Style World Prior and Presets}
\label{app:newworld}

This appendix details the game-world prior that informs our \emph{Type-Feature Generation} instantiation. It is a structured, lore-aware specification that our generator consumes as high-level constraints; no copyrighted maps or data are reproduced.

\textbf{Factions and territories:}
We assume three competing factions (green/purple/yellow) controlling contiguous administrative regions, plus several high-danger neutral zones in the far north. Every controllable region is paired with one settlement (safe hub) and one fort (conflict anchor). Neutral zones host outposts instead of player towns. This prior supplies (i) a faction histogram over the island, (ii) a per-region requirement of \{settlement, fort\}, and (iii) neutral-only constraints for the most dangerous areas.

\textbf{Biomes and regional diversity:}
Major biome families include temperate forest/grassland, wetlands, tropical/coastal, highlands/snow, arid desert, and special supernatural forests. The prior ties each territory template to a biome mix (e.g., \emph{WindswardLike}: plains/forest heavy; \emph{ShatteredMountainLike}: tundra/highland; \emph{EdengroveLike}: lush supernatural forest), guiding $N_{\texttt{type}}$ quantile mapping and post-hoc coastal/height adjustments (e.g., beaches at low elevation, snow at high elevation).

\textbf{Danger bands and difficulty gradient:}
The island is partitioned into four normalized danger bands: \texttt{Green} (entry), \texttt{Yellow} (intermediate), \texttt{Red} (high), and \texttt{Black} (endgame). Templates supply target proportions for these bands. Band membership conditions feature ranges (e.g., hp/dps multipliers) and rarity priors (ordinary $\gg$ elite $\gg$ boss), creating a spatially coherent difficulty ladder.

\textbf{Resources and gathering:}
Resource families (wood, stone, ore, fluids, alchemy/ammo, fibers/textiles, herbs/spices, crops/produce, fungi) are tied to biome masks and modulated by $N_{\texttt{feat}}$ for clustered richness. Higher tiers (e.g., rare trees or ores) are restricted to \texttt{Red}/\texttt{Black} bands and suitable biomes (e.g., mountainous or desert highlands). Quotas and per-class radii prevent over-saturation and enforce discoverable clusters rather than uniform scatter.

\textbf{Wildlife (neutral fauna):}
Non-hostile or conditionally hostile wildlife (e.g., turkey, rabbit, deer/elk, bison) spawn primarily in \texttt{Green}/\texttt{Yellow} bands, gated by biome suitability (open plains, forest edges, wetlands) and kept sparse enough to signal world “aliveness” without overwhelming tactical content. Intensities derive from dedicated $N_{\texttt{feat}}$ fields and are clipped by danger bands to avoid unrealistic concentration inside the harshest zones.

\textbf{Enemy families and rarity:}
Hostile families-Corrupted, Lost, Ancient, Angry Earth, Beast, and Humanoid-are tied to biomes, landmarks, and danger bands. The template defines which families dominate in which contexts (e.g., Corrupted in endgame mountains; Lost in wetlands and ruins; Ancient in relic sites; Angry Earth in enchanted forests). Rarity is staged by thresholds on intensity: ordinary enemies are widespread in \texttt{Green}/\texttt{Yellow}; elites concentrate in \texttt{Red}; bosses appear only at the top percentiles in \texttt{Red}/\texttt{Black}, often co-located with strong landmarks.

\textbf{Landmarks and travel:}
We place settlements (safe hubs), forts (conflict anchors), outposts (neutral rest stops), spirit shrines (fast travel), expedition entrances (dungeons), elite zones (outdoor group challenges), and miscellaneous quest hubs. Counts per class are template-defined; positions are chosen by constrained optimization that balances even coverage, biome/danger compatibility, and per-class spacing, with $N_{\texttt{feat}}$ providing saliency cues (e.g., “most prominent spot” within a territory).

\textbf{Presentation and outputs:}
Five-view bundles render semi-transparent territory layers plus DPI-aware iconography for points of interest. Legends and scale bars are standardized across runs. Optional GeoJSON includes, for each point: class/type, biome/danger band, intensity-derived feature values (e.g., elite/boss flags), enabling downstream analytics or coupling with simulation modules.

\textbf{Scope and originality:}
The prior is inspired by public descriptions of New World's world structure (factions, territory control, biomes, danger progression, content categories) but serves only as a thematic scaffold. All rasters, placements, and attributes are generated from our Perlin-based fields and template quotas; seeds guarantee reproducibility while allowing diverse, non-copied worlds.

\section{Experimental Setup Details}
\label{app:exp-details}

\subsection{Behavior Parameterization}
We evaluate how dual Perlin fields modulate large neutral crowds to achieve local spatial coherence without global lock-step. Experiments are conducted as a multi-scale batch and all reporting aggregates over 20 fixed seeds. Unless a setting is explicitly varied by the batch protocol, we fall back to the nominal defaults.

\textbf{Environment and multi-scale protocol:}
The world is a square torus of side $L{=}1000$ (meters), discrete time uses a unit step $\Delta t{=}1$ at $1\,\mathrm{Hz}$, and agents integrate
$\mathbf{x}_i(t{+}1)=\mathbf{x}_i(t)+v_i(t)\,[\cos\theta_i(t),\sin\theta_i(t)]$ with wraparound.
Three batch scales are run under matched logging (per-tick summaries and periodic snapshots):
\emph{small}: $N{=}200$, $T{=}360$, snapshots every $5$ ticks;
\emph{medium}: $N{=}1200$, $T{=}720$, snapshots every $10$ ticks;
\emph{large}: $N{=}3200$, $T{=}1080$, snapshots every $15$ ticks.
A fixed set of 20 seeds is used across methods to enable paired comparisons.
Two auxiliary sweeps complement the batch:
(i) a \emph{Perlin-scale} sweep with $(f_{\theta}, f_{v}, v_{\mathrm{drift}})\in\{(0.006,0.0065,0.001),(0.01,0.011,0.002),(0.016,\allowbreak0.017,0.004)\}$;
(ii) an \emph{agents-scale} sweep with $N\in\{500,1000,2000,\allowbreak4000\}$ (other settings fixed).

\textbf{Our Perlin instantiation:}
We use two independent multi-octave Perlin stacks (distinct seeds/frequencies) for headings and speeds with temporal \emph{drift} coherence ($v_{\mathrm{drift}}{=}0.002$ unless swept). The heading field uses $(f{=}0.01,\,K{=}4,\,p{=}0.5,\,\ell{=}2.0)$ with a small angular jitter ($0.02$); the speed field uses $(f{=}0.011,\,K{=}4,\,p{=}0.5,\,\ell{=}2.0)$ mapped to $v\in[v_{\min},v_{\max}]$ with $v_{\min}{=}0.6$, $v_{\max}{=}1.4$. Headings are blended by vector averaging with $\beta{=}0.9$ to avoid wrap artifacts. Speeds are smoothed by EMA with $\rho{=}0.8$ by default; an OU-like alternative
$v_i(t{+}1)=v_i(t)+\beta\,(v_i^\star{-}v_i(t))+\sigma\,\xi_t$
is available (e.g., $\beta{=}0.9$, $\sigma{=}0.05$).
When the multi-scale batch does not specify a value (e.g., jerk or spectrum windows), we use the nominal defaults given under \emph{Metrics} below.

\textbf{Baselines:}
We compare against six representative controls under matched duration, logging, and seed sets:
\emph{Perlin-single} (one shared scalar field drives both $\theta$ and $v$);
\emph{URW} (uncorrelated random walk with heading noise $\sigma_\theta{=}0.4$ and fixed speed);
\emph{OU-heading} \cite{uhlenbeck1930theory} (OU update on heading with $\beta{=}0.9$, $\sigma_\theta{=}0.1$);
\emph{Curl-noise} (divergence-free flow $\mathbf{v}\propto\nabla^\perp\psi$ from a Perlin potential with $f{=}0.01$, $v_{\mathrm{drift}}{=}0.002$);
\emph{Vicsek} \cite{vicsek1995novel} (neighbor alignment within radius $R{=}20$ under angular noise $\eta{=}0.25$ at constant $v{=}1.0$);
\emph{Piecewise} (coarse piecewise-constant vector grid (cell size $80$) with bilinear blending).

\textbf{Metrics:}
Spatial structure is summarized in distance bins by directional similarity
$S_{\mathrm{dir}}(r){=}\mathbb{E}[\cos(\Delta\theta)]$
and speed correlation $C_v(r)$, together with angular/speed semivariograms $\gamma_\theta(r)$ and $\gamma_v(r)$ and correlation lengths (distance at which the statistic decays to a preset fraction). The multi-scale batch uses extended bins
$[0,10,20,30,40,60,80,100,140,180,240,320,420,520]$.
Temporal smoothness includes short-lag autocorrelations of $\theta$ and $v$, jerk statistics with window $1$,
$\mathrm{jerk}(t){=}\|\Delta \mathbf{v}(t{+}1)-\Delta \mathbf{v}(t)\|$ (mean and $95$th percentile), and spectral high/low-frequency energy ratios computed on the kinetic signal $\sum_i v_i^2(t)$ (FFT window $256$).
Diversity covers polarization
$\Phi{=}\|\sum_i \hat{\mathbf{v}}_i\|/N$ (mean/std; lower indicates less alignment),
directional entropy $H(\theta)$, and speed distribution (mean/std/skew).
Coverage and path quality use a sliding window of $60$ ticks to compute recent visited-fraction and path tortuosity (mean/p95). To avoid ambiguity, we distinguish two notions of coverage in this paper: in the behavior/action studies we report coverage fraction, a 0–1 area-visit ratio; in the spawn studies we report coverage distance, the mean nearest-entity distance (in world units). These quantities have different semantics and units and should not be compared directly.
Efficiency reports per-tick runtime mean and method-specific noise samples per tick.
All metrics are computed per run and aggregated over 20 seeds for each method and scale.

\subsection{Activation Timing}

We evaluate temporal fields that modulate when and where activity concentrates under quotas and cooldowns. All studies follow a multi-scale batch protocol: unless explicitly overridden by the scale settings below, we use the nominal defaults; each configuration is replicated over fixed seed sets to enable paired comparisons. We consider two complementary experiments: (A) per-agent action starts driven by hazard/phase fields, and (B) spawn placement driven by space–time Perlin couplings.

\subsubsection{Perlin-driven Action Timing}

We test whether Perlin hazard/phase drivers produce locally synchronized yet globally de-locked action starts relative to stochastic, deterministic, and constraint-oriented baselines. We run fixed three-scale batches on a square torus with unit time step ($\Delta t{=}1$) and $1\,\mathrm{Hz}$ ticks, holding logging and analysis constant:
\begin{itemize}[noitemsep, topsep=0pt, leftmargin=*]
  \item Small: $N_{\text{agents}}{=}800$, map side $L{=}600$, horizon $T{=}1200$.
  \item Medium: $N_{\text{agents}}{=}2000$, map side $L{=}1000$, horizon $T{=}1800$.
  \item Large: $N_{\text{agents}}{=}8000$, map side $L{=}2000$, horizon $T{=}3600$.
\end{itemize}
Unless otherwise stated, we report the Medium scale as the baseline (warmup $60$, snapshots every $5$ ticks). For robustness, we additionally sweep Perlin hyperparameters with
$f\in\{0.005,0.01,0.02\}$, $K\in\{3,4,5\}$, $p\in\{0.45,0.55\}$, $\ell\in\{1.8,2.2\}$.

We instantiate two Perlin-based start drivers with optional convex mixing. A persistent Perlin grid (default $f{=}0.01$, $K{=}4$, $p{=}0.5$, $\ell{=}2.0$) is updated by slow drift ($v_{\mathrm{drift}}{=}0.002$) and sampled at agent positions.
\emph{Hazard variant (default):} agent-wise intensities are formed with base rate $\lambda_0$ and floor $\varepsilon$, smoothed by exponential averaging (e.g., $0.5$), and globally mean-normalized to the target rate specified in \S\ref{sec:method} (hazard-only unless the phase variant is explicitly enabled).
\emph{Phase variant (optional):} field values map to local phases (cycle length $T_{\mathrm{cycle}}{=}60$, jitter $1$), with a circular Gaussian kernel (bandwidth $8$) concentrating starts near each agent's phase; a convex mixing weight $\alpha$ interpolates hazard and phase when hybrid behavior is desired.

\textbf{Baselines:}
We compare against:
\emph{Poisson} (constant-rate Bernoulli with $\lambda$ matched to the Perlin mean);
\emph{Filtered random} (Poisson proposals with neighborhood radius $R_{\mathrm{sync}}{=}20$, cap $K{=}2$, window $w{=}2$ suppressing nearby starts);
\emph{Fixed} (periodic starts with period $\mathcal{N}(8,2^2)$, uniform initial phase, and tick-level jitter);
\emph{Constraint-token} (space partitioned into cells each with $K_{\text{cell}}{=}3$ capacity tokens, priority queuing, max wait $8$);
\emph{Constraint–round-robin} (grid of regions with $12$ slots/region, rotation period $30$ to balance concurrency);
\emph{Sinusoid} (global modulation $\lambda(t){=}\lambda_0[1{+}A\sin(2\pi t/P)]$ with $A{=}0.3$, $P{=}120$);
\emph{Hawkes-inhibitory} \cite{hawkes1971spectra} (self-inhibiting conditional intensity with negative excitation $\alpha{<}0$, exponential temporal kernel, and compact spatial kernel at $R_{\mathrm{inhib}}{=}20$; window $w{=}3$).

\textbf{Metrics:}
We quantify timing regularity by inter-event-interval statistics (mean/std/CV), duty cycle, inactivity-gap p95, burstiness $B{=}(\sigma{-}\mu)/(\sigma{+}\mu)$ and Fano factor \cite{fano1947ionization} $F{=}\mathrm{Var}(N_w)/\mathbb{E}[N_w]$ over sliding windows; temporal smoothness by second-difference energy and spectrum energy ratios of the active-count series; spatial balance by regional coefficient of variation (grid-based) and Moran's $I$ \cite{moran1950notes}; and space–time structure by \emph{front coherence} (correlation with the Perlin phase driver over the cycle) and second-order spatial coupling via Ripley’s $K(r)$ and the pair-correlation $g(r)$ 
(under CSR, $K(r)=\pi r^2$ and $g(r)=1$) \cite{ripley1976second, baddeley2016spatial}. We use two notions of temporal smoothness: ISI CV computed on the merged event-timestamp sequence, and spectral smoothness of the global active-count series (HF/LF). These measure different aspects and may rank methods differently. Operational cost is reported as per-tick runtime and decision throughput. All metrics are computed per run and aggregated over the seed set for each scale.

\subsubsection{Perlin-based Spawn Placement}

We assess two Perlin space–time couplings for proposing spawn candidates under replenishment quotas, cooldowns, eliminations, and player interactions. A non-toroidal square world is simulated with $T_{\mathrm{cycle}}{=}600$, $\Delta t{=}1$, and long horizons; the replenishment controller admits candidates to track target populations and per-cycle quotas. Multi-scale batches apply the following overrides:
\begin{itemize}[noitemsep, topsep=0pt, leftmargin=*]
  \item Small: $T{=}2400$; bounds $90{\times}90$; target pop $80$; cycle quota $48$; spawn cooldown per cycle $48$; players $6$; coverage samples $1024$.
  \item Medium: $T{=}3600$; bounds $120{\times}120$; target pop $128$; cycle quota $96$; spawn cooldown per cycle $96$; players $8$; coverage samples $2048$.
  \item Large: $T{=}7200$; bounds $240{\times}240$; target pop $256$; cycle quota $192$; spawn cooldown per cycle $192$; players $12$; snapshot period $120$; coverage samples $4096$; spatial radii $[5,10,20]$; temporal window $180$.
\end{itemize}
Unless scaled, default dynamics include monsters (speed $1.5$, persistence $0.85$, turn noise $0.5$, jitter $0.1$) and players (speed $2.4$, kill radius $1.75$, respawn delay $90$).

We now instantiate these ideas as two complementary policies.
\textbf{Perlin-A (space$\to$time; phase binning):} per cycle, a field $n(\mathbf{x})\!\in\![0,1]$ is regenerated; $M{=}256$ stratified candidates receive phases $\tau(\mathbf{x})=\lfloor n(\mathbf{x})T_{\mathrm{cycle}}\rfloor$; at tick $t$, proposals satisfy $t\bmod T_{\mathrm{cycle}}{=}\tau$, optionally thinned at rate $(1{-}\varepsilon)$ to reduce bursts. Typical settings use $(f{=}0.06, K{=}3, p{=}0.55, \ell{=}2.0, \varepsilon{=}0.1)$ and $128$ cycle sites.
\textbf{Perlin-B (time$\to$space; iso-band front):} the cycle angle $\theta$ maps to $\ell(\theta)\in[0,1]$, forming $S(\theta)=\{\mathbf{x}:\,|n(\mathbf{x})-\ell(\theta)|\le\varepsilon\}$; proposals are sampled uniformly from $S(\theta)$ or via Farthest-Point Sampling (greedy max–min distance; a.k.a.\ B-Far: $x^\star=\arg\max_{x\in S}\min_{q\in E}\|x-q\|_2$)~\cite{gonzalez1985clustering,mitchell1987generating,bridson2007fast}. A jittered time grid with minimum separation spaces iso-band evaluations across the cycle; typical defaults use spacing $\approx T_{\mathrm{cycle}}/20$, selection \emph{random}, and per-cycle counts controlled by $(\text{mean},\text{min},\text{max}){=}(3.0,1,6)$. A tuned variant uses tighter spacing (e.g., $15$), jitter $5$, min-separation $8$, slightly adjusted $(f,\ell,\varepsilon)$, and \emph{farthest} selection.

\textbf{Baselines:}
We benchmark against \emph{Uniform random} (UR), \emph{Filtered random} (safety/spacing filters), \emph{Poisson-disk} (per-cycle blue-noise sites), \emph{MVN+Poisson} (mixture-of-Gaussians density with Poisson counts), \emph{Facility-location} (distance- and desirability-aware greedy scoring with throttling), and \emph{Sinusoid} (global temporal modulation without spatial structure). All methods are rate matched by target population and per-cycle quotas.

\textbf{Metrics:}
Spatial coverage and balance are summarized by mean coverage distance (mean nearest-entity distance; reported in world units), regional coefficient of variation, Moran's $I$, average nearest-neighbor distance, and pair-correlation summaries. Temporal stability uses inter-spawn-interval CV, spectral smoothness, and \emph{front coherence} computed against the cycle-indexed time-to-level map. Operational statistics include per-tick runtime, spawns/eliminations per cycle, and load variance. Per-scale results are aggregated over fixed seed ranges, and comparisons emphasize coverage, regional balance, time stability, and efficiency under identical replenishment settings.

\subsection{Type-Feature Generation}
We instantiate Type-Feature Generation as a New World-style synthetic territory builder that decouples discrete layouts (factions, biomes, danger bands, content types) from continuous attributes (density multipliers, rarity, power scaling). The world is a square island of extent $8\,\mathrm{km}$ rendered at $300\,\mathrm{DPI}$ with a $1024{\times}1024$ base raster and a $512{\times}512$ parameter grid. All layers are driven by a seed bundle $(layout,noise,place)$; a user-visible integer seed $s$ is deterministically mapped to these substreams via fixed affine offsets to guarantee run-level reproducibility and per-layer independence. See Appendix~\ref{app:newworld} for the lore-aware world prior and presets that parameterize our generator.

\textbf{Environment and priors:}
We employ three territory templates (\emph{WindswardLike}, \emph{ShatteredMountainLike}, \emph{EdengroveLike}) to specify high-level mixes and quotas: (i) target histograms for faction/biome/danger bands; (ii) per-class quotas for resources, wildlife, enemy families, and landmarks; (iii) per-class exclusion and minimum-spacing radii; and (iv) feature ranges per danger band (e.g., hp/dps multipliers, elite/boss probabilities). The dual-field design uses low-frequency $N_{\texttt{type}}$ stacks for discrete maps (regional quantile mapping to hit mixes and maintain contiguity) and mid/high-frequency $N_{\texttt{feat}}$ stacks for continuous intensities (monotone transforms to target value ranges while preserving rank order). A height layer provides independent relief for coastal and ridge adjustments. Danger is sliced into four normalized bands with default thresholds $\texttt{Green}\le 0.35$, $\texttt{Yellow}\le 0.60$, $\texttt{Red}\le 0.82$, else $\texttt{Black}$.

\textbf{Noise layer specifications:}
Unless overridden by a template, the global noise configuration applies layer-specific multi-octave Perlin stacks:
\[
\begin{aligned}
&\text{faction: } f{=}0.018,\ K{=}4,\ p{=}0.45,\ \ell{=}2.1;\quad
\text{biome: } 0.022,\ 4,\ 0.50,\ 2.05;\\
&\text{danger: } 0.024,\ 5,\ 0.52,\ 2.15;\quad
\text{type: } 0.055,\ 6,\ 0.50,\ 2.20;\\
&\text{feature: } 0.110,\ 5,\ 0.55,\ 2.25;\quad
\text{height: } 0.028,\ 5,\ 0.48,\ 2.00.
\end{aligned}
\]
A small global drift (e.g., $v_{\mathrm{drift}}\approx 0.015$) can be enabled for cinematic phase shifts without breaking reproducibility.

\textbf{Placement and quotas:}
Given $\text{Map}_{\texttt{disc}}$ (faction/biome/danger/type rasters) and continuous feature fields, point sets for resources, wildlife, enemy families, and landmarks are sampled by inhomogeneous Poisson processes gated by the relevant intensity, clipped by masks, and filtered by per-class minimum distances. Quotas are enforced at category or regional granularity to match template budgets while avoiding local saturation. Elite/boss rates are conditioned on danger bands via band-specific feature ranges to produce the intended rarity gradient (ordinary $\rightarrow$ elite $\rightarrow$ boss).

\textbf{Rendering and exported artifacts:}
Each run produces a five-view bundle: (1) \emph{overview} (factions + danger outlines + landmarks + rare content), (2) \emph{resources\_biomes} (biome fills with resource clusters and rare markers), (3) \emph{enemies\_danger} (danger fills with enemy families/rarities), (4) \emph{landmarks\_admin} (faction fills with towns/forts/shrines/expeditions), and (5) \emph{zoom\_quadrants} (detail plates of resource peaks, enemy peaks, and landmark clusters). Optional vector exports include per-point attributes (type, danger band, intensity-derived features), enabling downstream analysis and integration. Legend layout, transparency, and symbol-density controls are kept constant across templates to support apples-to-apples comparisons.

\textbf{Instantiation used in our experiments:}
We showcase three seeds, \emph{WindswardLike} (seed 42), \emph{ShatteredMountainLike} (seed 7), and \emph{EdengroveLike} (seed 99), each rendered as the five-view bundle. Unless a template demands overrides, the layer stacks and danger thresholds above are used verbatim. Icon density and labeling adopt consistent presentation settings (e.g., moderate legend panel, symbol caps per $\mathrm{km}^2$) to keep map readability and per-class visibility stable across runs.

\section{Extended Experimental Results}
\label{app:extended-results}

This appendix provides the full numerical results, additional figures, and extended discussion for the three parameter–field families evaluated in Section~\ref{sec:results}. We retain the same experimental setup as in Section~\ref{sec:experiments} and organize the material by direction.

\subsection{Behavior Parameterization}
\label{app:extended-results-behavior}

\begin{figure}[t]
  \centering
  \includegraphics[width=\columnwidth]{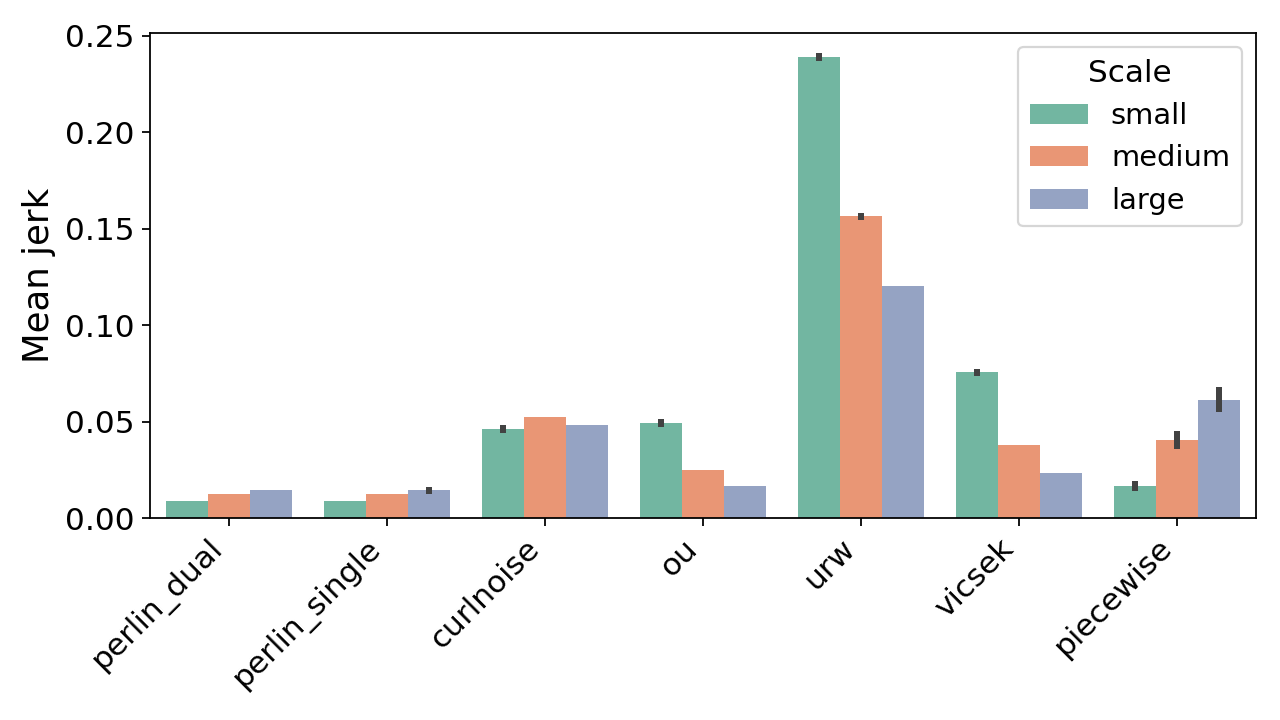}
  \caption{Mean jerk (lower better). Perlin achieves low jerk across scales.}
  \label{fig:bp_jerk}
\end{figure}

\begin{figure}[t]
  \centering
  \includegraphics[width=\columnwidth]{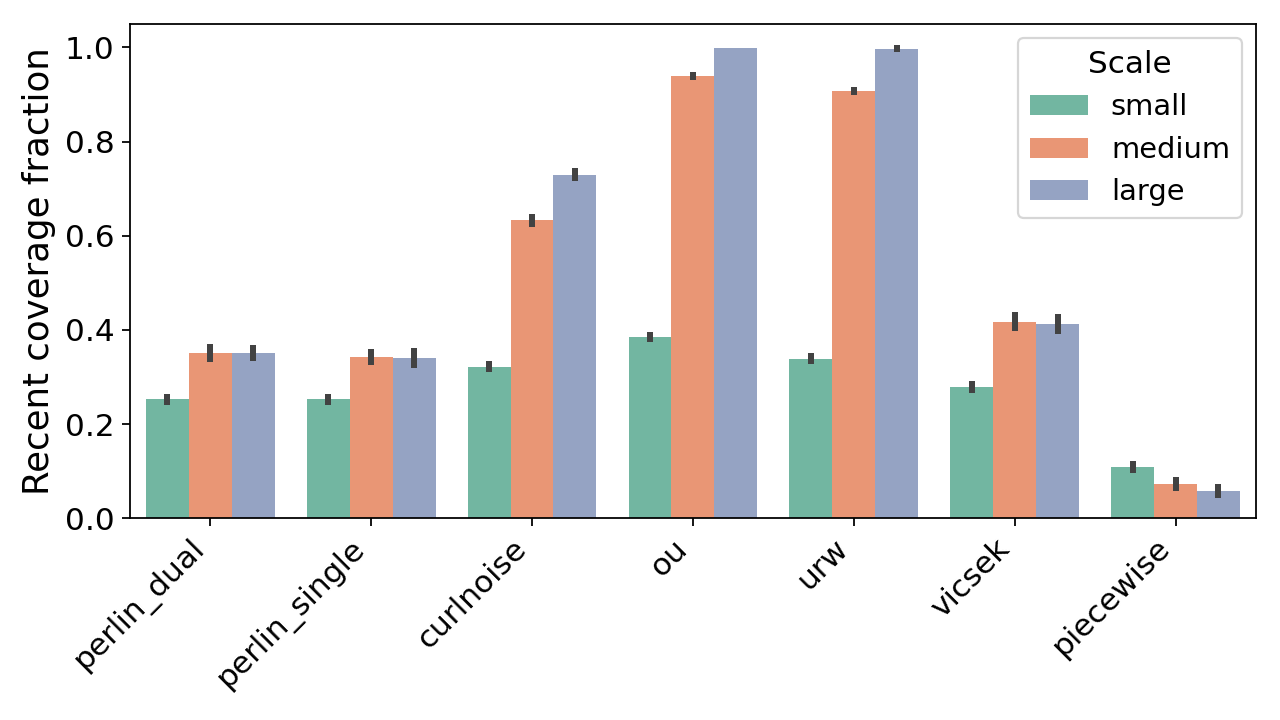}
  \caption{Recent coverage fraction (higher better).
  Perlin is moderate; \texttt{piecewise} is low; \texttt{ou}/\texttt{urw} are highest but unstructured.}
  \label{fig:bp_cov}
\end{figure}

\begin{table}[t]
\centering
\caption{Medium scale:
mean\(\pm\)95\%\,CI over seeds. Best (or tied) per column in \textbf{bold}.}
\label{tab:bp_medium}
\small
\begin{tabular}{lcccc}
\toprule
Method & \(S_{\mathrm{dir}}@5\uparrow\) & Jerk\(\downarrow\) & Cov.\(\uparrow\) & Runtime ms\(\downarrow\) \\
\midrule
perlin\_dual   & \textbf{0.989}\(\pm\)0.001 & \textbf{0.013}\(\pm\)0.000 & 0.350\(\pm\)0.013 & 9.88\(\pm\)1.43 \\
perlin\_single & \textbf{0.989}\(\pm\)0.001 & \textbf{0.013}\(\pm\)0.000 & 0.342\(\pm\)0.011 & 10.12\(\pm\)1.41 \\
curlnoise      & 0.869\(\pm\)0.006 & 0.052\(\pm\)0.000 & 0.633\(\pm\)0.007 & 12.18\(\pm\)1.63 \\
ou             & 0.007\(\pm\)0.011 & 0.024\(\pm\)0.000 & \textbf{0.939}\(\pm\)0.004 & \textbf{6.01}\(\pm\)0.86 \\
urw            & 0.004\(\pm\)0.011 & 0.156\(\pm\)0.000 & 0.907\(\pm\)0.003 & 6.07\(\pm\)0.90 \\
piecewise      & 0.773\(\pm\)0.031 & 0.041\(\pm\)0.003 & 0.074\(\pm\)0.008 & 6.39\(\pm\)0.85 \\
vicsek         & 0.977\(\pm\)0.000 & 0.038\(\pm\)0.000 & 0.417\(\pm\)0.015 & 375.45\(\pm\)54.16 \\
\bottomrule
\end{tabular}
\end{table}

Across three batch scales, dual Perlin fields have the strongest local spatial coherence and very smooth kinematics at lightweight runtime; exploration coverage is moderate (above piecewise but below random/OU), and global polarization is high—an expected trade-off for field–driven motion. Figure~\ref{fig:bp_sdir} in the main text shows that both \texttt{perlin\_dual} and \texttt{perlin\_single} attain \(S_{\mathrm{dir}}@5\) values near 1.0 at all scales (e.g., medium: \(0.989\pm0.001\)), exceeding \texttt{curlnoise} (\(\approx0.87\)) and far above stochastic baselines (\texttt{ou}/\texttt{urw} \(\approx 0\)). \texttt{vicsek} attains comparably high \(S_{\mathrm{dir}}@5\) but at a much higher runtime cost (Table~\ref{tab:bp_medium}).

Temporal smoothness is summarized in Figure~\ref{fig:bp_jerk}. Perlin methods are among the smoothest (medium jerk \(0.013\pm0.000\)), clearly below \texttt{curlnoise} and far below \texttt{urw}; \texttt{vicsek} is also smooth but computationally expensive. These jerk statistics are consistent with the short-lag autocorrelation and spectral HF/LF ratios (not shown), which indicate that Perlin concentrates energy at low frequencies and avoids high-frequency jitter in the kinetic signal.

Exploration and global alignment trade off as shown in Figure~\ref{fig:bp_tradeoff}. Perlin methods sit in the high-coherence / moderate-coverage / high-polarization regime (coverage $\sim 0.35$ at medium scale), outperforming \texttt{piecewise} on coverage while conceding coverage to \texttt{ou}/\texttt{urw}, which lack local coherence (near-zero \(S_{\mathrm{dir}}@5\)). Figure~\ref{fig:bp_cov} further reports recent coverage by method and scale, highlighting that Perlin remains in a stable mid-range and clearly exceeds \texttt{piecewise}, while the most diffusive baselines (\texttt{ou}/\texttt{urw}) maximize coverage at the expense of any recognizable flow structure.

Table~\ref{tab:bp_medium} summarizes key metrics at the Medium scale. Perlin methods dominate on local coherence and jerk, achieve reasonable coverage, and incur per-tick runtime (\(\approx 10\) ms) that is comparable to lightweight baselines and far below \texttt{vicsek}. Cross-scale robustness is strong: Perlin's \(S_{\mathrm{dir}}@5\) and jerk exhibit negligible degradation from small (200 agents) to large (3200 agents), and coverage rises slightly with scale for all methods without changing the relative ordering. The main limitation is elevated global polarization, reflecting field-aligned motion. In deployments where de-locking is paramount, polarization can be reduced by mixing multiple Perlin orientation fields, increasing drift or curl components, or injecting low-variance heading jitter, all of which are compatible with our framework and preserve local coherence.

\subsection{Activation Timing: Action Starts}
\label{app:extended-results-activation-action}

\begin{figure}[t]
  \centering
  \includegraphics[width=\columnwidth]{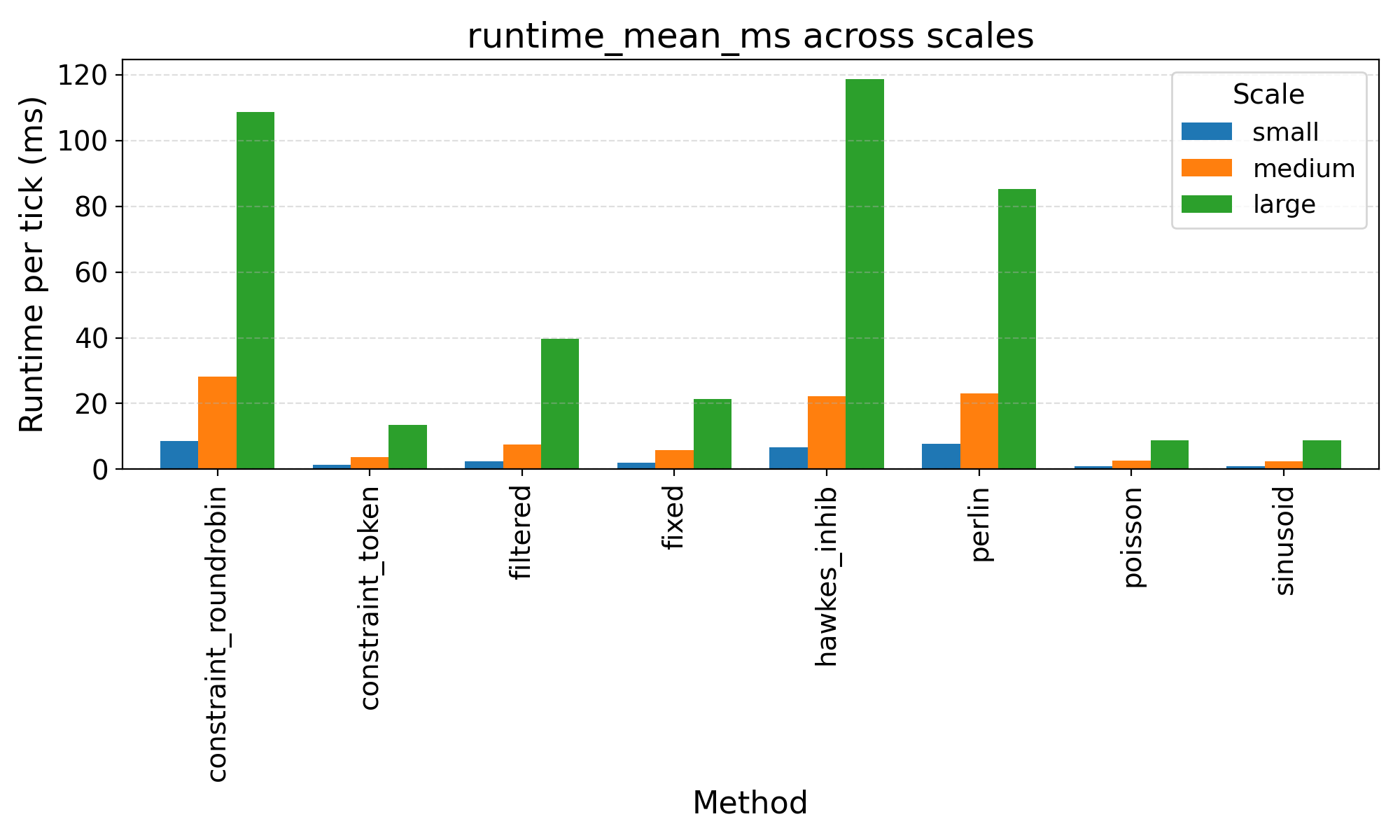}
  \caption{Per-tick runtime across scales. Perlin is moderate (grid sampling + smoothing), below capacity-heavy roundrobin at Large.}
  \label{fig:at_runtime}
\end{figure}

\begin{table}[t]
  \centering
  \setlength{\tabcolsep}{4pt}
  \footnotesize
  \begin{tabular}{lccccc}
    \toprule
    Method & Coverage~$\uparrow$ & Duty & HF/LF~$\downarrow$ & Fano~$\downarrow$ & Runtime~(ms)~$\downarrow$ \\
    \midrule
    Perlin & 0.879 & 0.106 & 0.122 & 0.924 & 23.1 \\
    Poisson & 0.895 & 0.113 & 0.119 & 0.954 & 2.6 \\
    Filtered & 0.924 & 0.126 & 0.133 & 0.910 & 7.5 \\
    Sinusoid & 0.886 & 0.112 & 0.015 & 4.874 & 2.4 \\
    Hawkes-inhib & 0.869 & 0.098 & 0.140 & 0.913 & 22.3 \\
    Constraint-Token & 0.895 & 0.113 & 0.124 & 0.945 & 3.8 \\
    Constraint-Roundrobin & 0.155 & 0.022 & 0.014 & 3.204 & 28.2 \\
    Fixed & 0.998 & 0.309 & 0.191 & 0.906 & 5.8 \\
    \bottomrule
  \end{tabular}
  \caption{Medium-scale metrics (means). Perlin pairs low HF/LF with near-Poisson burstiness and strong coverage at matched duty cycle; Sinusoid is ultra-smooth but globally synchronized (high Fano); capacity methods either underutilize (Roundrobin) or trade coverage for constraints; Fixed is period-driven rather than rate-matched.}
  \label{tab:at_medium}
\end{table}

The main text highlighted Perlin's ability to match target duty cycles and maintain strong coverage without inducing global waves. Here we expand on temporal statistics and runtime. In the rate-matched group (Perlin, Poisson, Filtered, Sinusoid, Hawkes-inhib, Constraint-Token), duty cycles align closely with the target across scales (Figure~\ref{fig:at_duty_cycle}), confirming that global mean-rate normalization of the Perlin hazard achieves precise rate control. Fixed is intentionally not rate-matched and runs at a higher duty, while Constraint-Roundrobin enforces quotas and therefore underutilizes capacity.

Temporal smoothness is measured by the high/low-frequency energy ratio (HF/LF) of the global active-count series. At the Medium scale, Perlin's HF/LF ($0.122$) is on par with Poisson ($0.119$) and lower than Filtered ($0.133$) and Hawkes-inhibitory ($0.140$); only Sinusoid and Constraint-Roundrobin yield lower HF/LF (around 0.015), but they do so via global synchronization and strict capacity ceilings. The Fano factor further shows that Perlin's fluctuations are close to Poisson-like (Fano $0.924$ vs.\ $0.954$), indicating stable variance without heavy-tailed surges. In contrast, Sinusoid exhibits a large Fano factor ($4.874$) due to its strong global oscillations, and Constraint-Roundrobin combines low HF/LF with low duty and high burstiness under token constraints.

Spatial coverage increases with scale for all methods (Figure~\ref{fig:at_coverage}). At Medium, Perlin's coverage ($0.879$) is competitive with Poisson ($0.895$) and Filtered ($0.924$) and clearly ahead of capacity-limited Roundrobin ($0.155$). At Large, Perlin saturates close to full coverage, showing that its space–time coherence does not harm exploration when agent densities are realistic. Table~\ref{tab:at_medium} aggregates key metrics at the Medium scale, highlighting that Perlin combines low HF/LF and near-Poisson Fano with strong coverage and matched duty cycle.

Runtime per tick scales roughly linearly with the number of agents (Figure~\ref{fig:at_runtime}). On our setup, Perlin incurs $\sim\!7.8$\,ms (Small), $23.1$\,ms (Medium), and $85.6$\,ms (Large) per tick, sitting above simple Poisson/Filtered/Sinusoid baselines but comparable to Hawkes-inhibitory and below constraint-heavy Roundrobin at Large scale. Given the improved temporal regularity and avoidance of global waves, this overhead is often acceptable in offline scheduling or server-side orchestration scenarios. The main caveats are that Perlin does not reach the extreme smoothness of a pure sinusoid (by design), and that under extreme external capacity constraints its benefits can be partially masked by enforced ceilings.

\subsection{Activation Timing: Spawn Placement}
\label{app:extended-results-activation-spawn}

\begin{figure*}[!t]
  \centering
  \includegraphics[width=\linewidth]{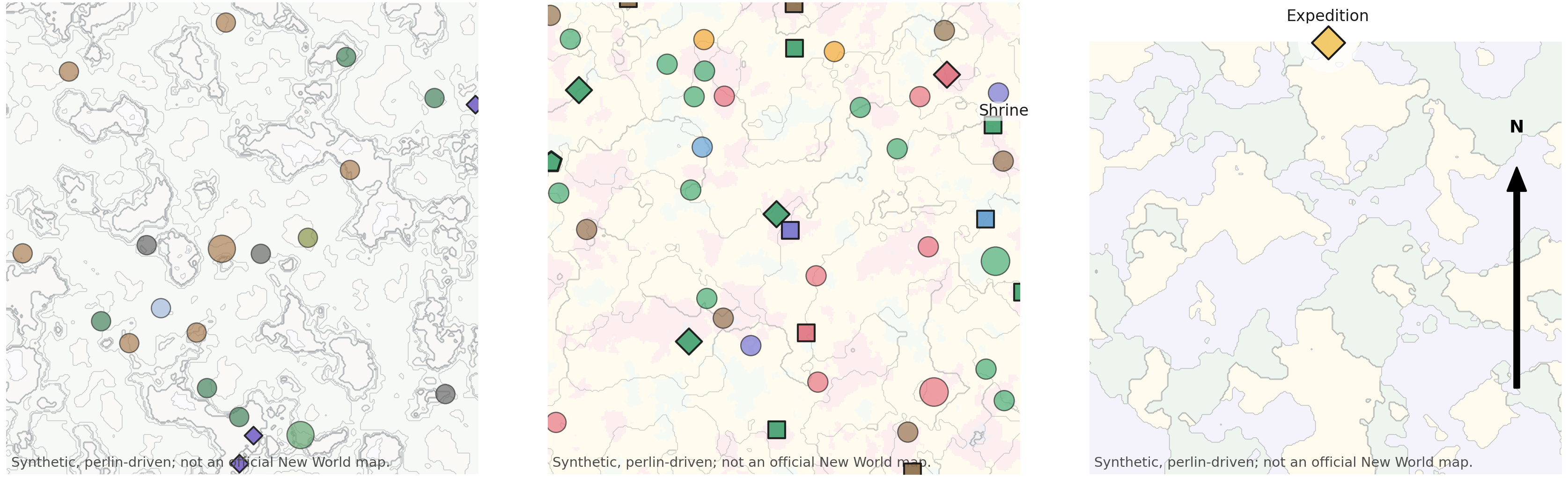}
  \caption{\textbf{Local detail (zoom quadrants) for \texttt{EdengroveLike}, seed 99.}
  Three zooms from the five–view bundle illustrate: \emph{left}—clustered resource pockets shaped by feature contours; \emph{middle}—a shrine placed at a corridor pinch with mixed biomes/danger levels; \emph{right}—an expedition entrance embedded in a highland enclave.}
  \label{fig:nw_zoom}
\end{figure*}

\begin{figure}[t]
  \centering
  \includegraphics[width=\linewidth]{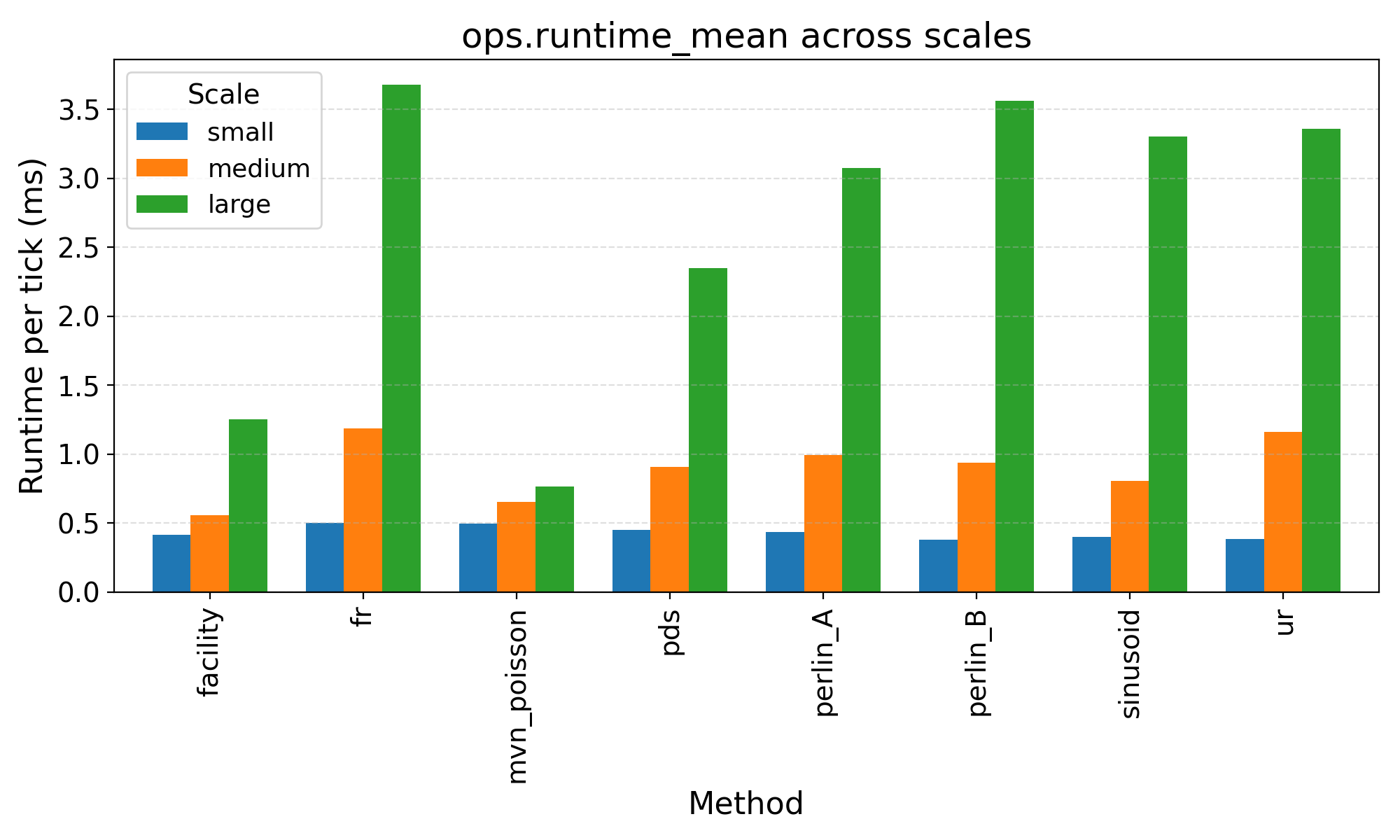}
  \caption{\textbf{Runtime per tick across scales (ms).}
  Perlin-A/B are in the middle of the pack with near-linear growth; MVN+Poisson and Facility are the cheapest; FR and PDS are somewhat higher at Large scale.}
  \label{fig:spawn-runtime}
\end{figure}

\begin{table}[t]
  \centering
  \caption{\textbf{Key metrics (Medium / Large).} Values are approximate means across seeds. Coher.\ = front coherence (higher is better); ISI CV = coefficient of variation of inter-spawn intervals (lower is smoother); Coverage = coverage distance (mean nearest-entity distance; world units; lower is better).}
  \begin{tabular}{lcccc}
    \toprule
    Method & Coher. & ISI CV & Coverage & Runtime \\
    \midrule
    Perlin-A & 0.011 / 0.010 & 1.23 / 1.58 & 3.03 / 4.38 & 1.00 / 3.07 \\
    Perlin-B & 0.010 / -0.003 & 1.39 / 1.42 & 3.63 / 4.81 & 0.94 / 3.56 \\
    FR       & 0.013 / 0.006  & 0.087 / 0.047   & 2.87 / 4.05 & 1.18 / 3.68 \\
    UR       & 0.002 / -0.002 & 0.36 / 0.03   & 2.52 / 3.51 & 1.16 / 3.36 \\
    PDS      & 0.0057 / 0.0018  & 3.96 / 4.49 & 2.56 / 2.53 & 0.91 / 2.35 \\
    Sinusoid & -0.003 / -0.0018 & 0.81 / 0.79 & 4.02 / 5.58 & 0.81 / 3.30 \\
    \bottomrule
  \end{tabular}
  \label{tab:spawn-medium-large}
\end{table}

The main text contrasted Perlin-A and Perlin-B with lightweight spatial baselines using phase-front coherence and coverage distance. Here we provide more detailed temporal and runtime statistics. As shown in Figure~\ref{fig:spawn-coherence}, Perlin-A maintains a small but positive coherence across all scales (approximately $0.024/0.011/0.010$ for small/medium/large), indicating a reproducible, gently moving front. Perlin-B's coherence hovers near zero ($-0.039/0.010/-0.003$), consistent with its design goal of balancing exposure rather than locking to a front. Some baselines, such as Facility, exhibit sign flips as the spatial pattern changes, making them less predictable when retargeted to new maps.

Temporal smoothness is summarized by the inter-spawn interval coefficient of variation (ISI CV) and load variance (not shown). Table~\ref{tab:spawn-medium-large} reports that Perlin-A and Perlin-B maintain ISI CV in the 1.2–1.6 range, markedly smoother than Poisson-disk sampling (3.96/4.49) and MVN+Poisson (5.53/2.31). Sinusoid exhibits low spectral HF/LF but does so via global waves. Filtered Random can achieve very low ISI CV at small/medium scales, but this depends strongly on aggregate rate; at large scales, Uniform Random's ISI CV becomes very small simply due to high event counts, without offering any spatial structure. In contrast, Perlin couplings deliver smoothness while encoding explicit space–time patterns.

Spatial coverage and balance follow the patterns in Figure~\ref{fig:spawn-coverage}. At large scale, Perlin-A/B achieve coverage distances in the same range as UR ($\approx 3.51$) and PDS ($\approx 2.53$), indicating that the added space-time structure does not substantially degrade coverage distance under our settings. Filtered Random tends to attain slightly lower distances at medium and large scales, but at the cost of additional filtering overhead. MVN+Poisson reaches very large coverage values, but these reflect a clustering objective (placing entities in a few high-density hot spots) rather than balanced coverage; as a result, coverage distance is not directly comparable there and correlates with higher burstiness and uneven exposure.

Runtime per tick for the spawn experiment is shown in Figure~\ref{fig:spawn-runtime}. Perlin-A and Perlin-B lie in the middle of the pack and scale nearly linearly with the number of entities and candidate sites: representative means are $0.42/1.00/3.07$\,ms and $0.37/0.93/3.56$\,ms for small/medium/large, respectively. Filtered Random becomes somewhat more expensive at large scale ($\sim 3.7$\,ms), whereas Facility and MVN+Poisson remain the cheapest (around 1.2\,ms and 0.75\,ms at large). Thus, Perlin couplings provide a favourable balance of coverage, smoothness, and predictable runtime, especially when a structured yet de-locked space–time pattern is desired.

The main limitations are that Perlin-B's phase coherence remains near zero unless explicitly tuned for sharper phase focusing, and that Perlin methods do not maximize every coverage metric when compared to cluster-oriented objectives such as MVN+Poisson. Moreover, relying on a single coherence score can under-represent richer spatial pattern differences; this is why we pair it with coverage, ISI CV, and load diagnostics

\subsection{Type-Feature Generation for Synthetic Worlds}
\label{app:extended-results-world}

The zoom quadrants in Figure~\ref{fig:nw_zoom} complement the global views in Figure~\ref{fig:nw_bundle} by showing how the same field stack governs local micro-structure. In the left zoom, resource clusters follow gently meandering feature isocontours; high-intensity strips of the feature field translate into dense pockets of nodes, while low-intensity patches remain sparsely populated. The middle zoom shows a shrine located at a corridor pinch where biomes and danger bands mix, demonstrating how minimum-distance constraints and band-aware quotas together select sites that are both accessible and strategically useful. The right zoom highlights an expedition entrance embedded in a highland enclave: the entrance is surrounded by higher-danger territory with enough clear space for maneuvering, reflecting the template's requirements for elite/boss content.

Quantitatively, we observe small quota errors for faction/biome/danger histograms (typically within a few percentage points of the template targets) and nearest-neighbor spacing distributions that closely track the specified minimum radii for resources and enemies (full tables omitted for brevity). Co-location rates between enemy families and biomes, as well as between rare resources and high-danger bands, confirm the intended affinities: for example, the majority of elite/boss enemies spawn in Red/Black bands, while most rare nodes occur in mid-to-high danger tiers and appropriate biomes. These audits support the qualitative assessment in the main text that the dual-field design successfully decouples discrete layout control from continuous feature modulation.

There are, however, visible limitations. In very sparse danger bands, strict minimum-distance constraints can lead to slightly over-regular patterns, suggesting that additional jitter or multi-radius mixing would improve realism. Our current height field is generic and does not yet encode hand-authored vistas, road networks, or water bodies; integrating such constraints would further align the generator with production-grade worldbuilding. Finally, our evaluation focuses on static snapshots; incorporating dynamic processes such as resource depletion, re-growth, and NPC ecology would provide a richer test of how field-driven worlds behave over long-running simulations.

Overall, the extended results here reinforce the main paper's conclusions: low-frequency Perlin type fields, regional quantile mapping, and mid/high-frequency feature fields, combined with inhomogeneous Poisson placement and spacing rules, yield territories that are internally consistent, template-faithful, and reproducible, while still exhibiting natural variation across seeds and templates.


\end{document}